\documentclass[10pt,twocolumn,letterpaper]{article}

\usepackage{cvpr}              

\definecolor{cvprblue}{rgb}{0.21,0.49,0.74}
\usepackage{graphicx}
\usepackage{amsmath}
\usepackage{mathtools}
\usepackage{amssymb}
\usepackage{amsfonts}
\usepackage{booktabs}
\usepackage{color}
\usepackage{multirow}
\usepackage{subcaption}
\usepackage{caption}

\usepackage[accsupp]{axessibility} 

\usepackage[pagebackref,breaklinks=true,colorlinks,citecolor=blue,urlcolor=blue,linkcolor=blue,bookmarks=false]{hyperref}

\def\paperID{3964} 
\def\confName{CVPR}
\def\confYear{2025}

\title{DiffSensei: Bridging Multi-Modal LLMs and Diffusion Models \\ for Customized Manga Generation}

\author{
    Jianzong Wu$^{1,2}$ \quad
    Chao Tang$^1$ \quad
    Jingbo Wang$^2$ \quad
    Yanhong Zeng$^2$ \quad
    Yunhai Tong$^{1}$ \quad
    Xiangtai Li$^{3, 4}$ \\
   {\normalsize $^1$ Peking University \quad $^2$ Shanghai AI Laboratory \quad $^3$ Nanyang Technological University \quad $^4$ Bytedance Seed } \\
    {\normalsize Project Page: \url{https://jianzongwu.github.io/projects/diffsensei/}} \\
    {\normalsize \textit{Email: jzwu@stu.pku.edu.cn, xiangtai94@gmail.com}}
}

\begin{document}
\twocolumn[{%
\renewcommand\twocolumn[1][]{#1}%
\maketitle
\centering
\vspace{-5mm}
\includegraphics[width=1.0\linewidth]{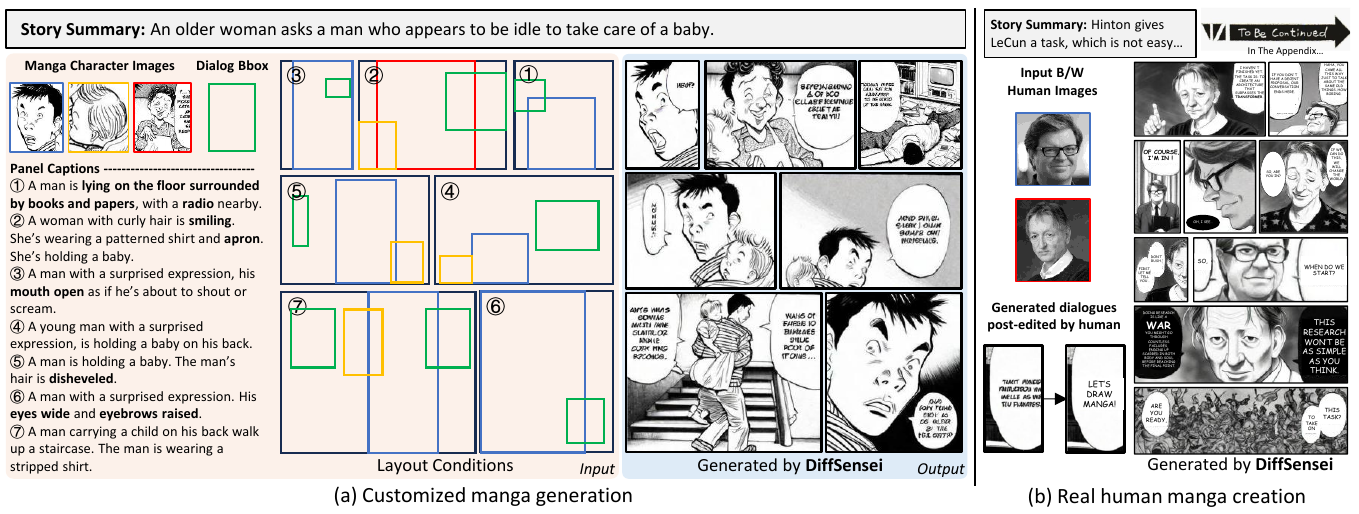}
\captionof{figure}{Results of \textbf{DiffSensei}. (a) Customized manga generation with controllable character images, panel captions, and layout conditions. Our DiffSensei successfully generates detailed character expressions and states following the panel captions. (b) Manga creation for real human images. The dialogues are post-edited by humans. The continuation is in the Appendix. We \textbf{strongly} recommend that the readers see the Appendix for more comprehensive results. Manga reading order: Right to left. Top to bottom.}
\label{fig:teaser}
\vspace{5mm}
}]

\begin{abstract}
Story visualization, the task of creating visual narratives from textual descriptions, has seen progress with text-to-image generation models. However, these models often lack effective control over character appearances and interactions, particularly in multi-character scenes.
To address these limitations, we propose a new task: \textbf{customized manga generation} and introduce \textbf{DiffSensei}, an innovative framework specifically designed for generating manga with dynamic multi-character control.
DiffSensei integrates a diffusion-based image generator with a multimodal large language model (MLLM) that acts as a text-compatible identity adapter. Our approach employs masked cross-attention to seamlessly incorporate character features, enabling precise layout control without direct pixel transfer.
Additionally, the MLLM-based adapter adjusts character features to align with panel-specific text cues, allowing flexible adjustments in character expressions, poses, and actions.
We also introduce \textbf{MangaZero}, a large-scale dataset tailored to this task, containing 43,264 manga pages and 427,147 annotated panels, supporting the visualization of varied character interactions and movements across sequential frames.
Extensive experiments demonstrate that   DiffSensei outperforms existing models, marking a significant advancement in manga generation by enabling text-adaptable character customization.
The code, model, and dataset are open-sourced to the community.
\footnote{The work is done in Shanghai AI Laboratory. Corresponding to: Xiangtai Li, Jingbo Wang.}
\vspace{-5mm}
\end{abstract}
\section{Introduction}
\label{sec:intro}
Story visualization, the process of generating visual narratives from textual descriptions, is a rapidly evolving field~\cite{ar-ldm, storygan, storygen, storydiffusion, seed-story, autostory, autostudio}. 
Among its various applications, manga generation holds particular significance due to its popularity and unique narrative requirements. Unlike traditional story visualization, manga demands consistent characters across panels, precise layout control for positioning multiple characters, and the seamless integration of dialogues in a coherent, visually engaging manner.

Currently, manga generation remains an underexplored area. Most existing research focuses on low-level image-to-image tasks, primarily converting general images to a manga style~\cite{sketch2manga, manga-background, manga-mimicking, mangagan}. While these tasks enhance the visual appeal of static images, they do not extend to generating fully customized manga content from scratch. In general story visualization, current approaches have demonstrated some success in generating coherent image sequences from text. Still, they cannot often customize the characters across scenes~\cite{ar-ldm, storygen, seed-story, storydiffusion}, a critical requirement for manga generation. Additionally, they do not provide the necessary control over layouts and dialogue placements, which are also essential for manga. A likely reason for these limitations is that existing story visualization datasets typically lack character annotations and layout controls~\cite{storygan, storygen, seed-story, VIST}.
Another research direction has explored zero-shot character customization, showing promise for achieving character customization across manga panels~\cite{ip-adapter, ms-diffusion, dreambooth, mix-of-show, instantid, anydoor, photomaker}. However, these approaches often result in rigid "copy-pasting" effects~\cite{anydoor, ms-diffusion, ip-adapter}, which limit expressive character variation and detract from narrative depth. This limitation largely stems from the scarcity of datasets capturing multiple appearances of the same character in varied expressions and poses.

To address these limitations, we propose a new task: \textbf{customized manga generation}. As illustrated in~\cref{fig:teaser}, this task focuses on creating manga images with multiple characters, each customized based on a character image and positioned according to user input. Characters must dynamically adapt to text prompts, altering their expressions, motions, and poses as the narrative unfolds. Dialog layout should also be managed to generate expressive manga panels. Compared to traditional story visualization tasks, this proposed task prioritizes essential manga-specific controllability, aiming to generate vivid manga panels that are both coherent and visually engaging while supporting the customization of multiple characters.
To address the lack of a dedicated dataset for customized manga generation, we collect a dataset of Japanese black-and-white manga, forming the basis of the proposed task. The resulting dataset, \textbf{MangaZero}, is the first large-scale collection designed to support multi-character, multi-state manga generation.

To tackle this task, we introduce a novel framework, \textbf{DiffSensei}, which leverages a diffusion-based image generator to produce customized manga panels. However, we observe that, even with training on multiple appearances of the same character, the generated portraits often tend to rigidly follow the pixel distribution of the input character image, resulting in limited variations in appearance, pose, and motion based on the text input.
Inspired by recent advancements in image editing using Multimodal Large Language Models (MLLMs)~\cite{lgvi, magicbrush, seed-story, seed-x, emu2, emu3, mgie, genartist, smartedit, wu2023open}, we propose using an MLLM as a text-compatible character adapter. This approach enables seamless, dynamic adjustments to characters in response to textual cues, thereby supporting coherent and expressive manga panel generation. Additionally, we incorporate masked attention injection to manage character layout, along with a dialog embedding technique tailored specifically for manga, allowing for precise control over dialog placement.
Through extensive experimentation, we validate DiffSensei's capability to generate coherent, expressive manga panels that maintain narrative consistency and offer improved character control. This represents a significant advancement in story visualization. 


\begin{itemize}
    \item We introduce a new task: \textbf{customized manga generation}, focused on generating manga images with multiple characters, each dynamically adapting to text prompts and positioned according to layout specifications.
    \item We present \textbf{MangaZero}, the first large-scale dataset specifically designed for multi-character, multi-state manga generation, addressing a significant gap in story visualization training data. The dataset will be released for the image generation community.
    \item We propose \textbf{DiffSensei}. As far as we know, it is the first framework for customized manga generation that links diffusion models and MLLMs. The MLLM serves as an adaptable character feature adapter, enabling characters to respond dynamically to textual cues. Extensive experiments demonstrate the effectiveness of DiffSensei.
\end{itemize}
\begin{table*}[t]
    \centering
    \caption{\textbf{Comparison between MangaZero and related publically available datasets.} A story is defined as a sequence of continuous images annotated consistently with character IDs. In MangaZero, a story means a manga page. A panel means a distinct story image, or called frame~\cite{seed-story, storygen}. Most series in MangaZero are still popular in 2024. Please see the Appendix for the dataset details.}
    \label{tab:dataset-compare}
    \scalebox{0.95}{
    \begin{tabular}{lccccccccccccc}
\toprule
\multirow{2}{*}{Dataset} & \multirow{2}{*}{Type} & \multirow{2}{*}{Resolution} & \multirow{2}{*}{\#Series} & \multirow{2}{*}{\#Stories} & \multirow{2}{*}{\#Panels} & \multicolumn{3}{c}{Annotations} & \multirow{2}{*}{Origin} \\
& & & & & & Caption & Character & Dialog & \\
\midrule
PororoSV~\cite{storygan} & Animation & Fix & 1 & 15,336 & 73,665 & \checkmark & $\times$ & $\times$ & 2003-2016 \\
FlintstonesSV~\cite{flintstones} & Animation & Fix & 1 & 25,184 & 122,560 & \checkmark & \checkmark & $\times$ & 1960-1966 \\
StorySalon~\cite{storygen} & Animation & Fix & 446 & 18,255 & 159,778 & \checkmark & $\times$ & $\times$ & YouTube \\
StoryStream~\cite{seed-story} & Animation & Fix & 3 & 12,614 & 257,850 & \checkmark & $\times$ & $\times$ & 1939-2013 \\
Manga109~\cite{manga109} & B/W Manga & Vary & 109 & 10,602 & 103,850 & $\times$ & \checkmark & \checkmark & 1970-2010 \\
\midrule
\textbf{MangaZero} & B/W Manga & Vary & 48 & \textbf{43,264} & \textbf{427,147} & \checkmark & \checkmark & \checkmark & 1974-2024 \\
\bottomrule
     \end{tabular}}
\end{table*}

\section{Related Work}

\noindent
\textbf{Story visualization.}
Story visualization, the process of generating visual narratives based on given stories, is rapidly evolving. Many approaches can generate consistent image sequences derived from story content~\cite{storydiffusion, autostory, ar-ldm, storygen, autostudio, seed-story}. Despite recent advancements, the field faces significant limitations. Most existing methods generate story images solely from text and image-level prompts~\cite{storydiffusion, ar-ldm, storygen, seed-story}, which restricts control over individual characters. This limited control over characters reduces the flexibility and depth of story visualization. A key factor is that current training datasets~\cite{VIST, storygan, flintstones, storygen, seed-story} lack character-specific annotations.
In response to the data limit, recent works~\cite{autostory, autostudio} explore multi-character control using training-free methods that leverage existing subject preservation techniques, such as IP-Adapter~\cite{ip-adapter}. Other works~\cite{textual-inversion, dreambooth, mix-of-show, ip-adapter, anydoor, ms-diffusion, instantid, photomaker, motionbooth, han2024face, han2023generalist, relationbooth} try to train diffusion models for a multi-character customized generation. However, these approaches often result in a "copy-pasting" effect, restricting the diversity of expressions and actions needed for dynamic storytelling. For training-free methods, combining multiple models can significantly slow down inference speeds.
To address these challenges, we first introduce a large-scale manga generation dataset with finely curated character annotations and then develop a novel framework utilizing diffusion models and MLLMs that enables the dynamic generation of manga panels.


\noindent
\textbf{Manga generation.}
The field of black-and-while manga generation has received limited exploration. Most existing researches focus on low-level image-to-image tasks, primarily transferring general images to a manga style~\cite{sketch2manga, manga-background, manga-mimicking, mangagan}. 
Recent works contribute to manga content understanding~\cite{magi, magiv2, manga-understanding-survey}.
In contrast, we propose the customized manga generation task beyond style transfer to offer complete character and story-driven manga generation.

\noindent
\textbf{MLLMs for personalized image generation.}
MLLMs have shown remarkable potential in personalized image generation, particularly for tasks involving image editing and customization~\cite{lgvi, magicbrush, seed-story, seed-x, emu2, emu3, mgie, genartist, smartedit, bai2024meissonic}. Notably, CAFE~\cite{cafe} explores customizing subject appearances through textual instructions. However, MLLM-driven image generation for multi-character narratives remains an open challenge, primarily due to the difficulties in maintaining inter-character relationships and scene continuity.
Our framework proposes an MLLM-based identity adapter that enhances dynamic story consistency in multi-character manga generation. In contrast to previous works, our framework takes multi-character features as input and edits these features collectively, following the text prompt, enabling flexible subject editing across multiple characters.


\section{The MangaZero Dataset}

In this section, we first define the problem in~\cref{sec:dataset-problem-formulation}. Then, we introduce the dataset construction pipeline in~\cref{sec:dataset-construction}.

\subsection{Problem Formulation}
\label{sec:dataset-problem-formulation}

We introduce a challenging new task: customized manga generation. This task focuses on generating manga images where multiple characters, each with their distinct image inputs, are customized and positioned by users. Importantly, characters must adapt to the text prompts by modifying their expressions, motions, and poses dynamically, even when only a limited set of character images is available.
To generate a manga story across $N$ panels (or frames), the inputs include: text prompts for each panel ${T_0, T_1, \ldots, T_{N-1}}$, character images $\mathbf{I} = {I_0, I_1, \ldots, I_{K-1}}$, character bounding boxes for each panel ${\mathbf{B}^c_0, \mathbf{B}^c_1, \ldots, \mathbf{B}^c_{N-1}}$, and dialogue bounding boxes for each panel ${\mathbf{B}^d_0, \mathbf{B}^d_1, \ldots, \mathbf{B}^d_{N-1}}$. The visualization of a panel is represented as $P_i = \Phi_\theta(T_i, \mathbf{I}, \mathbf{B}^c_i, \mathbf{B}^d_i)$, where $\Phi$ is the overall model function and $\theta$ represents the model's learned parameters.

\noindent
\textbf{Discussion.} 
This task diverges from existing story visualization and continuation tasks~\cite{ar-ldm, storygen}. Specifically, in story visualization tasks, a panel is produced using $P_i = \Phi_\theta(T_i)$, while in story continuation tasks, the panel generation depends on previous panels as $P_i = \Phi_\theta(T_i, T_{i-1}, P_{i-1})$ for $i > 0$. Both lack explicit character control, a crucial element in storytelling. Furthermore, the proposed task differs from subject-driven image generation approaches~\cite{dreambooth, ip-adapter, ms-diffusion}, as it demands that models not only generate accurate character representations but also modify characters’ attributes in response to panel captions and layouts, resulting in varied and coherent narrative visuals. Our experiments, detailed in~\cref{sec:exp}, demonstrate that our model significantly surpasses baseline models in these critical aspects.

\subsection{Dataset Construction}
\label{sec:dataset-construction}

In this section, we introduce the proposed large-scale manga story visualization dataset \textbf{MangaZero}.

%

\noindent
\textbf{Comparison with related datasets.}
A comprehensive comparison with existing datasets is presented in~\cref{tab:dataset-compare}. In contrast to current manga and story visualization datasets, the proposed MangaZero dataset stands out as being \textbf{larger in size}, \textbf{newer in source}, \textbf{richer in annotations}, \textbf{diverse in manga series}, and \textbf{varied in panel resolutions}. Compared to the well-known black-and-white manga dataset Manga109~\cite{manga109}, the MangaZero dataset encompasses more manga series published after the year 2000, which inspired its naming. Additionally, MangaZero includes famous series from before the year 2000 that are not featured in Manga109, such as Doraemon (1974).

\begin{figure}[t!]
    \centering
    \includegraphics[width=0.9\linewidth]{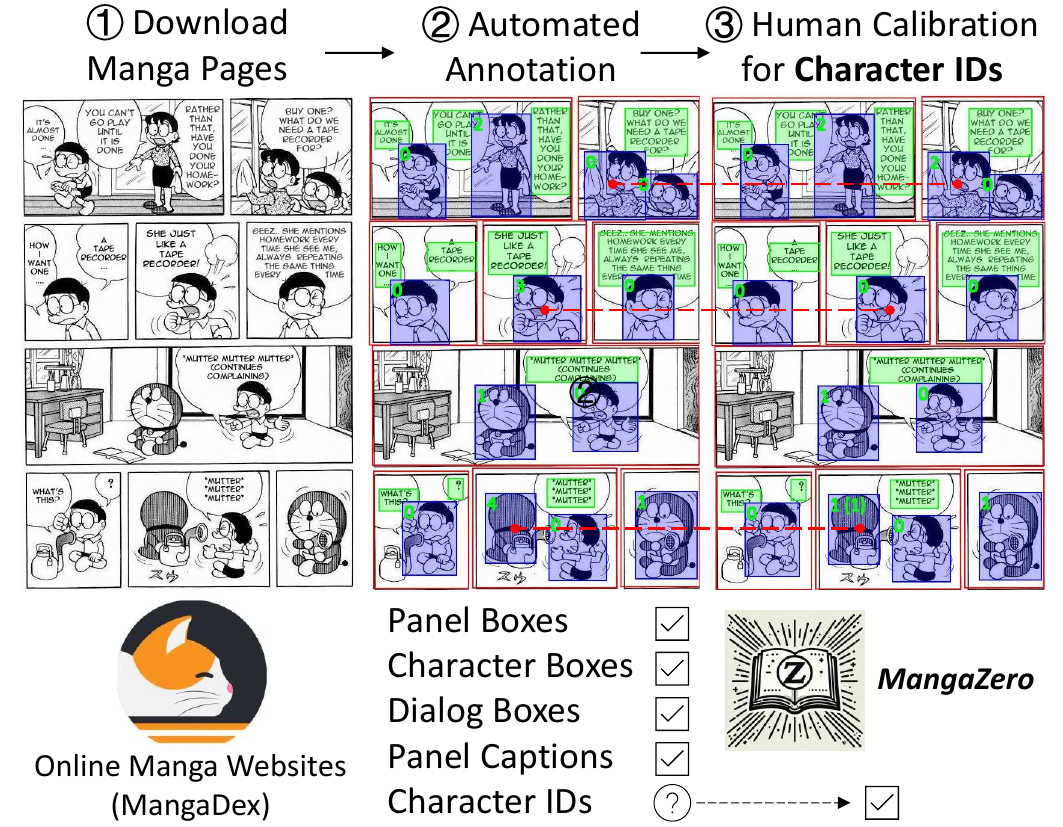}
    \caption{We construct \textbf{MangaZero} through three steps: 1) Download manga pages from the internet. 2) Annotate manga panels autonomously with pre-trained models. 3) Human calibration for the character ID annotation.}
    \label{fig:dataset-construction}
\end{figure}

\noindent
\textbf{Construction pipeline.}
To build our dataset, we first download manga pages from the internet, explicitly sourcing images from MangaDex~\cite{mangadex}. It is important to note that all data will be used solely for academic research, not commercial purposes. We select 48 manga series and download up to 1,000 pages per series, resulting in 43,264 double-page images.
These images are then annotated using pre-trained models. For manga-specific annotations, including panel bounding boxes, character bounding boxes, character IDs, and dialog bounding boxes, we employ the recent manga understanding model, Magi~\cite{magi}. It should be noted that character ID labeling is consistent only within individual pages, which is sufficient for achieving coherent character cross-reference.
Once the panel bounding boxes are obtained, we utilize LLaVA-v1.6-34B~\cite{llava} to generate captions for each panel. However, we observe that character ID labeling has relatively low accuracy, which poses a significant challenge for training purposes. To address this, human annotators refine the machine-generated labels, resulting in accurate and clean annotations. Finally, we split 96 pages (2 for each series) as the evaluation set and the remaining 43,168 pages as the training set.

\begin{figure*}[t!]
	\centering
	\includegraphics[width=1.0\linewidth]{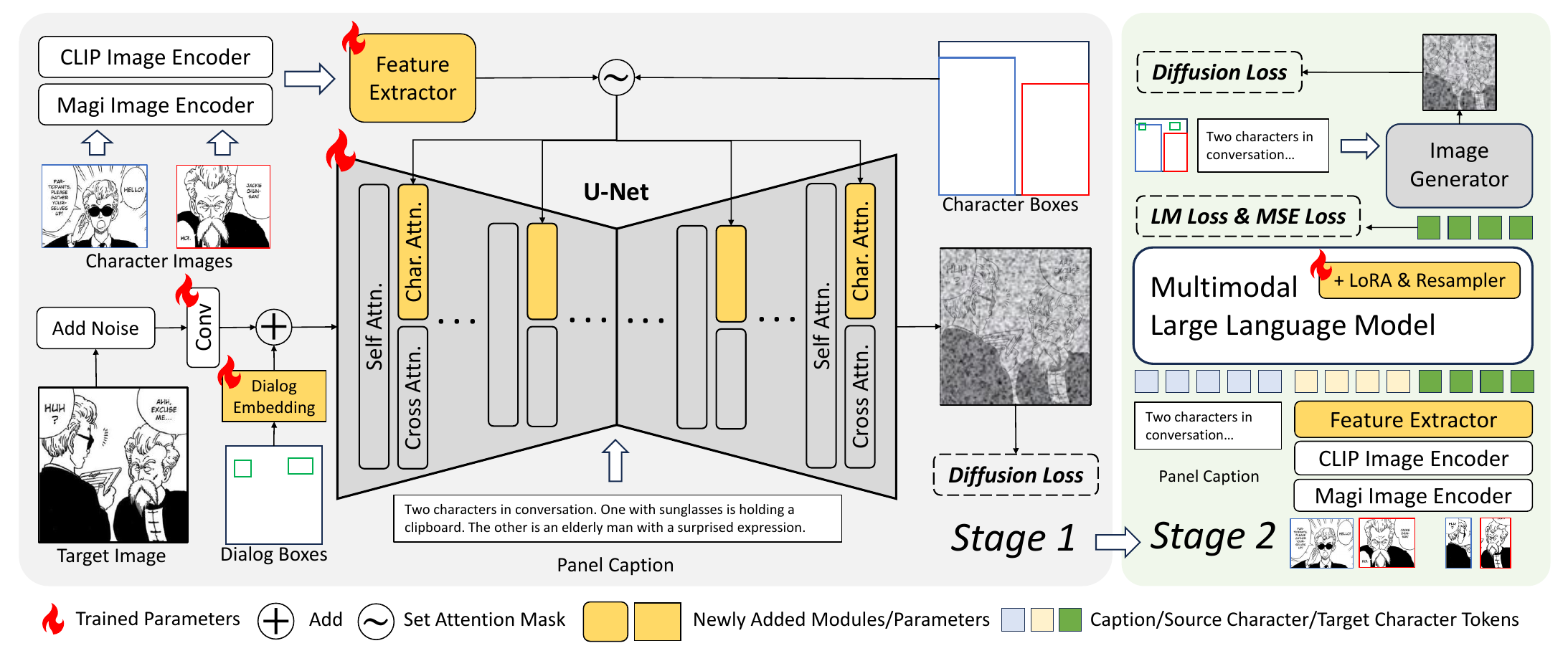}
	\caption{\textbf{The architecture of DiffSensei}. In the first stage, we train a multi-character customized manga image generation model with layout control. The dialog embedding is added to the noised latent after the first convolution layer. All the parameters in the U-Net and feature extractor are trained. In the second stage, we finetune LoRA and resampler weights of an MLLM to adapt the source character features corresponding to the text prompt. We use the model in the first stage as the image generator and freeze its weights.}
	\label{fig:architecture}
\end{figure*}

\section{Method}
\label{sec:method}

In this section, we introduce the architecture of the proposed framework, \textbf{DiffSensei}, which generates vivid manga panels with precise character and dialog layout control while adapting the characters' status flexibly.

\noindent
\textbf{Motivation.}
There are two critical problems in customizing objects and layouts in image generation: 1) Preserving the subjects' intrinsic features while avoiding direct copy-paste from source images, and 2) Ensuring reliable layout control with minimal computational cost during both training and inference.
To avoid copy-pasting effects, our model converts character image features into tokens, preventing the direct transfer of fine-grained pixel details. Additionally, we integrate an MLLM as a character image feature adapter. The MLLM adapter receives source character features and panel captions as inputs, generating text-compatible target character features. Compared with previous customization work~\cite{ms-diffusion}, this approach enables text-compatible character encoding and flexible character adaptation to captions. For layout control, we employ lightweight masked encoding techniques for both character and dialog layouts, significantly reducing computational costs compared with previous works~\cite{boxdiff, instancediffusion} while maintaining high accuracy in both training and inference phases. Experiment results in~\cref{sec:exp} demonstrate the effectiveness of our design.

\noindent
\textbf{Multi-character feature extraction.}
As illustrated in~\cref{fig:architecture}, we initially extract local image features using CLIP and image-level features from a manga image encoder. These two sets of features are then processed by a feature extractor, which is implemented as a resampler module~\cite{ip-adapter}. This process can be formalized as follows:
\begin{equation}
    \begin{aligned}
        \mathbf{c}_i &= \text{Resampler}([\text{CLIP}(\mathbf{I}), \psi(\mathbf{I})], \mathbf{q}, \mathbf{q}_{void}), \\
    \end{aligned}
\end{equation}
where $\psi$ represents the manga image encoder. $\mathbf{q}$ and $\mathbf{q}_{void}$ are trainable query vectors for character and non-character features, respectively. $\mathbf{q}$ re-samples the image features into the U-Net's cross-attention dimension, while $\mathbf{q}_{void}$ guides the cross-attention in regions without characters in the layout. $\mathbf{c}_i\in \mathbb{R}^{B\times ((N_c + 1)\times N_q)\times C}$ is the output feature for all characters, where $B$ is the batch size, $N_c$ is the maximum number of characters per panel (padded with all-zero features as needed), $N_q$ is the number of query tokens per character, and $C$ is the cross-attention dimension of the U-Net.
Through compressing the character images into a few tokens, DiffSensei avoids encoding fine-grained spatial features from reference images into the model~\cite{controlnet, storygen}. This enables focusing on the characters' semantic representations rather than rigid pixel distributions.

\noindent
\textbf{Masked cross-attention injection.}
%
We replicate the key and value matrices in each cross-attention layer, creating separate character cross-attention layers. 
This allows the image query features to attend to text and character cross-attentions independently and combine the results from both attentions. In the character cross-attention, we apply a masked cross-attention injection mechanism to control the layout of each character. 
Here, each character feature only attends to query features within its designated bounding box region. In areas without characters, query features attend to a placeholder vector, $\textbf{q}_{void}$. This can be formulated as:
\begin{equation}
    \begin{aligned}
        \hat{\mathbf{z}} &= \text{Softmax} \left( \frac{\mathbf{Q}\mathbf{K}_t^\top}{\sqrt{d}}\right) \mathbf{V}_t + \\
        & \alpha \cdot \text{Softmax} \left( \frac{\mathbf{Q}\mathbf{K}_i^\top}{\sqrt{d}} + \mathbf{M} \right) \mathbf{V}_i, \\
    \end{aligned}
\end{equation}
where $\mathbf{Q} = \mathbf{z}\mathbf{W}_q$, $\mathbf{K}^t = \mathbf{c}_t\mathbf{W}^t_k$, $\mathbf{V}^t = \mathbf{c}_t\mathbf{W}^t_v$, $\mathbf{K}^i = \mathbf{c}_i\mathbf{W}^i_k$, $\mathbf{V}^i = \mathbf{c}_i\mathbf{W}^i_v$. $\mathbf{Q}$ is the query, $\mathbf{W_q}, \mathbf{W}^t_k, \mathbf{W}^t_v$ are query, key, and value projection matrices for the text cross-attention. $\mathbf{W}^t_k$, $\mathbf{W}^t_v$ are key and value projection matrices for the character cross-attention, initialized from $\mathbf{W}^t_k$ and $\mathbf{W}^t_v$. $d$ is the key dimension. $\mathbf{c}_t$, $\mathbf{c}_i$ are text and character features respectively. $\mathbf{z}$, $\hat{\mathbf{z}}$ are the input and output image features. $\alpha$ is a hyperparameter that controls character attention weight. $\mathbf{M}$ is an attention mask to manage the characters' layout. Its values are defined as follows:
\begin{equation}
    \mathbf{M}[i, j] = 
    \begin{cases}
        0, & \text{if } j = N_c \text{ and } i \notin \mathbf{B}^c \text{ or } \\
        & \quad\quad\quad i \in \mathbf{B}^c[j] \\ 
        -\infty, & \text{otherwise}
    \end{cases}
\end{equation}
where $i$ denotes the position of query tokens, $j \in \{0, 1, ..., N_c\}$ is the character indices. The $N_c$-th character feature represents the placeholder vector $\mathbf{q}_{void}$. $\mathbf{B}^c[j]$ is the bounding box of the $j$-th character. The masked attention injection mechanism ensures that each character attends only to its specified bounding box region, while areas without characters attend to the placeholder vector. This technique achieves efficient and precise layout control for each character with minimal computational overhead.

\noindent
\textbf{Dialog layout encoding.}
Panels with dialog are a distinctive feature of manga images. However, most current text-to-image models struggle to generate coherent, readable text~\cite{stable-diffusion, sdxl}. While some recent models can produce stable text, they remain limited in terms of text length~\cite{flux}. Generating extended text, such as dialogues, continues to pose challenges.
Therefore, we propose controlling the layout of dialogs rather than the content itself. In this approach, human artists can manually edit the text within dialog bubbles, leaving image generation to the models.
Concretely, we introduce a trainable embedding to represent the dialog layout. The dialog embedding is first expanded to match the spatial shape of the noised latent and then masked with the dialog layout. By summing the masked dialog embedding with the noised latent, we can encode dialog positions within the image generator. This process is expressed as:
\begin{equation}
    \begin{aligned}
        \hat{\mathbf{z}}_t &= \text{Conv}(\mathbf{z}_t) + \text{Expand}(\mathbf{e}_d, \mathbf{z}_t)\cdot \mathbf{M}_d,
    \end{aligned}
\end{equation}
where $\mathbf{e}_d$ is the trainable dialog embedding, $\mathbf{z}_t$ is the noised latent in time step $t$, $\text{Expand}$ is a function that expands $\mathbf{e}_d$ to the latent shape, and $\mathbf{M}_d$ is the dialog region mask derived from input dialog bounding boxes $\mathbf{B}^d$. The output, $\hat{\mathbf{z}}_t$, serves as a dialog-layout-aware latent representation. This is then input into the U-Net for noise prediction. The dialog embedding effectively encodes the dialog layout, adding minimal computational overhead in space and time.

\noindent
\textbf{MLLM as text-compatible character feature adapter.}
After training the image generator, our model can effectively create manga panels that adhere to specified character appearances and layout conditions. However, the model often rigidly follows the input character images, lacking flexibility in adjusting expressions, poses, or motions based on panel captions.
We propose incorporating MLLM as a text-compatible character feature adapter. This approach allows dynamic modifications to character states based on text prompts. A training sample for MLLM is organized as [panel caption, source character image features, target character image features]. The image features are encapsulated by two special tokens, \textless IMG\textgreater and \textless/IMG\textgreater. To achieve this, we compute Language Modeling (LM) Loss on the special tokens to constrain the output format and Mean Squared Error (MSE) Loss to guide the target character features based on the panel caption. To ensure that the edited character features align with the image generator, we further pass the generated features into U-Net’s character cross-attention and compute diffusion loss. In this stage, only the LoRA and resampler weights in MLLM are updated. This process can be formalized as follows:
\begin{equation}
    \begin{aligned}
        \hat{\mathbf{h}}, \hat{\mathbf{c}}_i &= \text{MLLM}(T, \phi(\mathbf{c}_i)), \\
        \mathcal{L}_{lm} &= \text{LMLoss}(\hat{\mathbf{h}}, T), \\
        \mathcal{L}_{mse} &= \text{MSELoss}(\phi'(\hat{\mathbf{c}}_i), \tilde{\mathbf{c}}_i), \\
        \mathcal{L}_{diff} &= \mathbb{E}_{\mathbf{\epsilon}, t} || \mathbf{\epsilon} - \mathbf{\epsilon}_\theta(\mathbf{z}_t, T, \hat{\mathbf{c}}_i, \mathbf{B}^c, \mathbf{B}^d, t) ||^2,
    \end{aligned}
\end{equation}
where $T$ is the text prompt. $\phi$ and $\phi'$ denote the input and output resamplers of MLLM, which consist of stacked attention layers to convert the embeddings between inner and outer dimensions. $\hat{\mathbf{h}}$ refers to the MLLM predicted special token embeddings. We calculate LM Loss on it. $\hat{\mathbf{c}}_i$ is the predicted character features. We compute MSE Loss between $\hat{\mathbf{c}}_i$ and $\tilde{\mathbf{c}}_i$, the ground truth target character embedding extracted from the feature extractor. By leveraging the character ID annotations of MangaZero, we can obtain features from the same character across different panels, thus facilitating the training of the MLLM feature adapter. The adapted character feature $\hat{\mathbf{c}}_i$ is then passed to the image generator $\mathbf{\epsilon}$, previously trained, to compute a diffusion loss. The total loss for training MLLM is expressed as follows:
\begin{equation}
    \begin{aligned}
        \mathcal{L} &= \lambda_{lm}\mathcal{L}_{lm} + \lambda_{mse}\mathcal{L}_{mse} + \lambda_{diff}\mathcal{L}_{diff},
    \end{aligned}
\end{equation}
where $\lambda_{lm}$, $\lambda_{mse}$, and $\lambda_{diff}$ are loss weights.
\section{Experiments}
\label{sec:exp}

In this section, we thoroughly evaluate DiffSensei and compare it with baseline models.

\begin{table*}[!ht]
    \centering
    \caption{\textbf{Quantitative comparisons on automatic metrics}. Methods followed by ``*'' use reference images as input rather than characters. Methods marked by ``\dag'' means re-trained with dialog embedding.}
    \begin{subtable}{0.49\linewidth}
        \centering
        \caption{Comparison on MangaZero evaluation set.}
        \label{tab:exp-comparison-quantitative-mangadex}
        \scalebox{0.72}{\begin{tabular}{l|ccccc}
            \toprule
            Method & FID $\downarrow$ & CLIP $\uparrow$ & DINO-I $\uparrow$ & DINO-C $\uparrow$ & F1 score $\uparrow$ \\
            \midrule
            AR-LDM*~\cite{ar-ldm} & 0.409 & \textbf{0.257} & 0.548 & 0.507 & 0.004 \\
            StoryGen*~\cite{storygen} & 0.411 & 0.219 & 0.536 & 0.488 & 0.012 \\
            SEED-Story*~\cite{seed-story} & 0.411 & 0.169 & 0.416 & 0.405 & 0.006 \\
            StoryDiffusion*~\cite{storydiffusion} & 0.409 & 0.244 & 0.461 & 0.362 & 0.002 \\
            \midrule
            MS-Diffusion\textsuperscript{\dag}~\cite{ms-diffusion} & 0.408 & 0.229 & 0.610 & 0.641 & 0.720 \\
            \midrule
            \textbf{DiffSensei} & \textbf{0.407} & 0.235 & \textbf{0.618} & \textbf{0.651} & \textbf{0.727} \\
            \bottomrule        \end{tabular}}
    \end{subtable}
    \begin{subtable}{0.49\linewidth}
        \centering
        \caption{Comparison on Manga109 evaluation set.}
        \label{tab:exp-comparison-quantitative-manga109}
        \scalebox{0.72}{\begin{tabular}{l|ccccc}
            \toprule
            Method & FID $\downarrow$ & CLIP $\uparrow$ & DINO-I $\uparrow$ & DINO-C $\uparrow$ & F1 score $\uparrow$ \\
            \midrule
            AR-LDM*~\cite{ar-ldm} & 0.410 & \textbf{0.254} & 0.527 & 0.491 & 0.005 \\
            StoryGen*~\cite{storygen} & 0.414 & 0.214 & 0.540 & 0.493 & 0.004 \\
            SEED-Story*~\cite{seed-story} & 0.413 & 0.167 & 0.442 & 0.428 & 0.005 \\
            StoryDiffusion*~\cite{storydiffusion} & 0.410 & 0.238 & 0.442 & 0.355 & 0.001 \\
            \midrule
            MS-Diffusion\textsuperscript{\dag}~\cite{ms-diffusion} & \textbf{0.410} & 0.227 & 0.584 & \textbf{0.600} & 0.601 \\
            \midrule
            \textbf{DiffSensei} & \textbf{0.410} & 0.237 & \textbf{0.588} & \textbf{0.600} & \textbf{0.648} \\
            \bottomrule
        \end{tabular}}
    \end{subtable}
\end{table*}

\begin{figure*}[t!]
	\centering
	\includegraphics[width=1.0\linewidth]{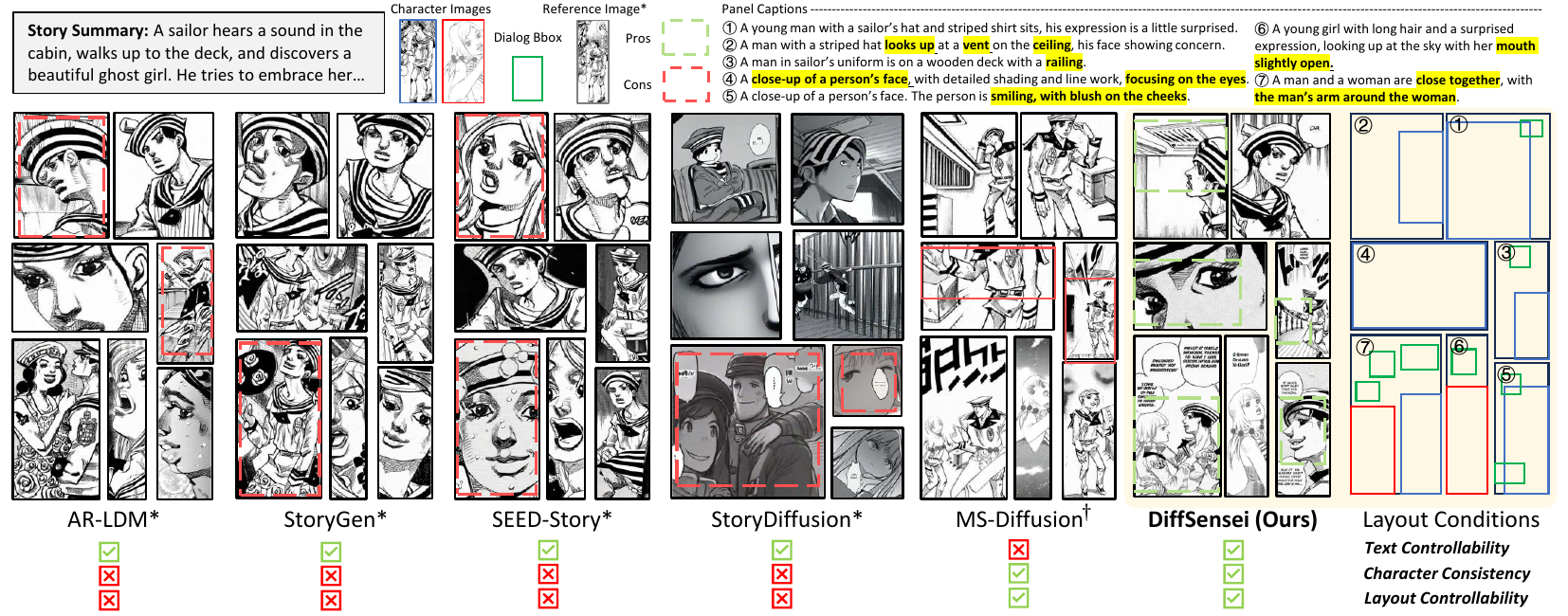}
	\caption{\textbf{Qualitative comparison with baselines}. Baselines followed by a ``*'' use reference images as input rather than character images. Methods marked by ``\dag'' means re-trained with dialog embedding. Our model excels at preserving the characters while following the text prompt. Our DiffSensei successively generates highlighted details in panel captions. Better viewed with zoom-in.}
	\label{fig:exp-comparison-qualitative}
\end{figure*}

\subsection{Experimental Settings}

\noindent
\textbf{Implementation details.}
The image generator is constructed on top of SDXL~\cite{sdxl}. The feature extractor's weights are initialized using the pre-trained IP-Adapter-Plus-SDXL~\cite{ip-adapter}, while the MLLM (Multi-modal Large Language Model) is initialized from SEED-X~\cite{seed-x}. Other newly introduced parameters, including the LoRA and resampler weights of the MLLM, are initialized randomly.
In stage 1, the image generator is optimized using a learning rate of 1e-5. Stage 2 training employs a learning rate of 1e-4 and a LoRA rank of 64~\cite{lora}. The optimizer is AdamW~\cite{adamw}. The loss function coefficients $\lambda_{lm}$, $\lambda_{mse}$, and $\lambda_{diff}$ are set to 1.0, 6.0, and 1.0, respectively. We train 250k steps for the first stage and 20k for the second stage. The source character images are chosen randomly, with a 50\% probability of being from the same page; otherwise, they are selected from the target image.
To handle varying image resolutions during training, we adopt the bucket-based approach from prior works~\cite{sdxl}, grouping images into resolution-specific buckets. For each training batch, images are loaded from the same resolution bucket. The batch size varies between 8 and 64 in stage 1, and between 8 and 128 in stage 2. This dynamic batch sizing is necessary to prevent out-of-memory (OOM) issues, especially when processing large-resolution images.
Please see more details in the Appendix.

\noindent
\textbf{Evaluation datasets and metrics.}
We evaluate our model using the MangaZero and Manga109~\cite{manga109} evaluation sets. Note that the model is only trained on MangaZero. Characters in Manga109 are \textbf{unseen} during training, serving as a benchmark for generalization.
To assess the generation quality of individual images, we employ autonomous metrics, which include Fréchet Inception Distance score (FID)~\cite{fid}, CLIP image-text similarity (CLIP)~\cite{clip}, DINO image similarity (DINO-I)~\cite{dinov2}, DINO character image similarity (DINO-C), and the dialog bounding box F1 score (F1 score). 
The source character images are randomly sampled on the same page. The dialog bounding boxes in the generated images are predicted using Magi~\cite{magi}.
For evaluating the story visualization quality of image sequences, human preference study proves more effective. We recruit human volunteers to choose their preferred story pages from our model’s output and baseline models in the MangaZero evaluation set. The evaluation criteria include five key aspects: text-image alignment, style consistency, character consistency, image quality, and overall preference.

\noindent
\textbf{Baselines.}
We select recent advanced story visualization models as our baselines, including StoryDiffusion~\cite{storydiffusion}, AR-LDM~\cite{ar-ldm}, StoryGen~\cite{storygen}, SEED-Story~\cite{seed-story}, and MS-Diffusion~\cite{ms-diffusion}. StoryDiffusion~\cite{storydiffusion} is a training-free method. We directly use an SDXL text-to-image model finetuned on MangaZero for evaluation. Despite that, we re-train other baselines on our dataset for a fair comparison.

\begin{figure}[t!]
	\centering
	\includegraphics[width=1.0\linewidth]{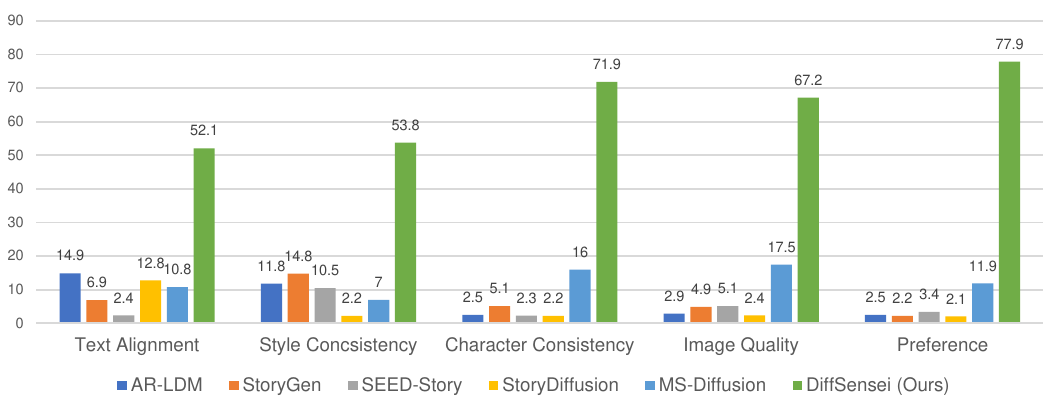}
	\caption{\textbf{Human preference study on MangaZero eval set}.}
	\label{fig:exp-human-preference}
    \vspace{-5mm}
\end{figure}

\begin{figure*}[t!]
	\centering
	\includegraphics[width=1.0\linewidth]{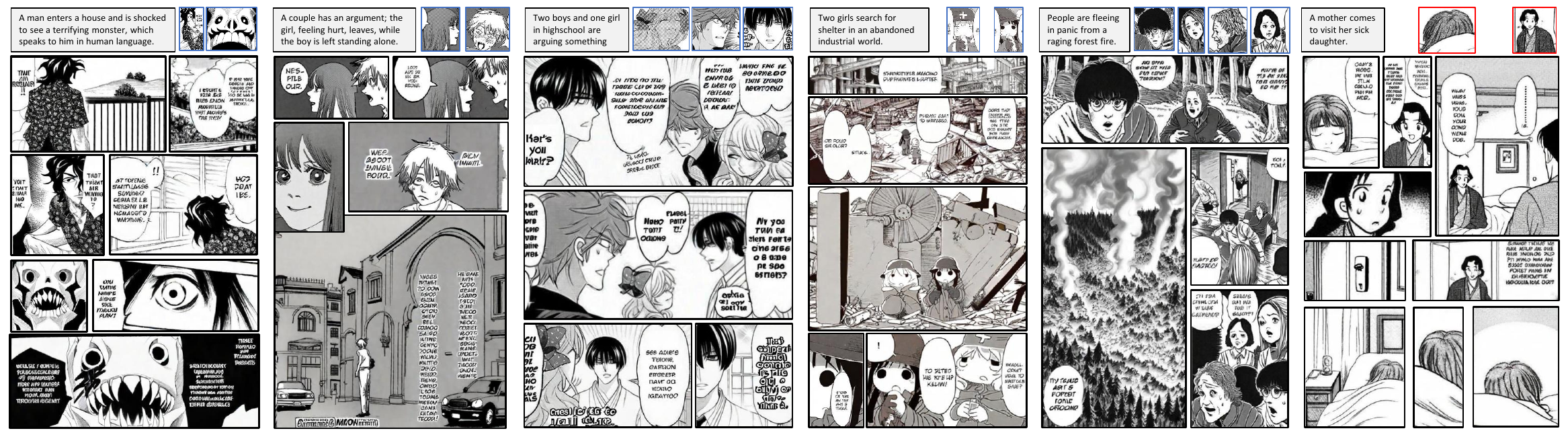}
	\caption{\textbf{Qualitative results}. Character images in red boxes are from Manga109 (The rightmost example). Our DiffSensei can generate vivid manga pages in various scenarios. Better viewed with zoom-in. More results can be found in the appendix.}
	\label{fig:exp-qualitative-results}
\end{figure*}

\subsection{Comparison to Baselines}

\noindent
\textbf{Quantitative comparison.}
We quantitatively compare our DiffSensei model with baseline models using automatic evaluation metrics. 
The results on the MangaZero evaluation set are presented in~\cref{tab:exp-comparison-quantitative-mangadex}. 
The results highlight that DiffSensei consistently outperforms the baseline models across five key metrics. Our model improves 0.06 in the CLIP metrics compared to the multi-subject customization baseline, MS-Diffusion~\cite{ms-diffusion}, which struggles to modify characters' states effectively in response to textual prompts. Furthermore, DiffSensei demonstrates superior image quality and character preservation, as evidenced by higher DINO-I and DINO-C scores. Although AR-LDM~\cite{ar-ldm} achieves higher CLIP metrics, it suffers from poor image quality and lacks the architectural capability to manage multiple characters, resulting in low DINO-C scores. In contrast, our method strikes a balance between maintaining character appearances and adapting to text prompts.
We also compare the Manga109 evaluation set with the results shown in~\cref{tab:exp-comparison-quantitative-manga109}. When using previously unseen characters as inputs, our model continues to outperform the baselines. These results underscore the strong generalization ability of DiffSensei, demonstrating effective adaptation to new characters.

In~\cref{fig:exp-human-preference}, we present the results of a human preference study comparing our model to the baselines across several dimensions. Our model receives the highest ratings from human evaluators, particularly in terms of overall preference, character consistency, and image quality. These findings confirm that DiffSensei excels in rendering vivid and engaging manga stories.

\noindent
\textbf{Qualitative comparison.}
\cref{fig:exp-comparison-qualitative} shows a qualitative comparison between DiffSensei and baseline models. The results illustrate that our model significantly outperforms the baselines in generating an entire page of a manga story. SEED-Story~\cite{seed-story} employs an MLLM to create captions for each panel, leading to unnatural narrative text and chaotic story generation that fails to form a coherent story. StoryDiffusion~\cite{storydiffusion} is limited to producing fixed-resolution images due to its self-attention sharing mechanism, restricting its ability to generate diverse images. It shows inferior results, probably because the input reference panel has an unbalanced aspect ratio. 
MS-Diffusion~\cite{ms-diffusion} trains with source character images from the target panel and lacks the flexibility to modify characters' states effectively. 
In contrast, our method excels in text-following, character preservation, and overall story presentation.

\subsection{Qualitative Results}

\cref{fig:exp-qualitative-results} shows several manga pages generated by DiffSensei. The results demonstrate that our method can generate vivid manga panels and provide visually plausible results for the customized manga generation task. Our model can also generalize to unseen characters, as illustrated in the rightmost example, with character images from Manga109~\cite{manga109} as input. 
Please see more results in the Appendix.

\subsection{Ablation Study}

\begin{table}[!t]
    \centering
    \caption{\textbf{Ablation study}. CM is character masked attention injection. DM is dialog masked encoding. Magi means using Magi~\cite{magi} image encoder. MLLM means using MLLM for stage 2 training.}
    \label{tab:exp-ablation}
    \scalebox{0.63}{
    \begin{tabular}{cccc|ccccc}
\toprule
CM & DM & Magi & MLLM & FID $\downarrow$ & CLIP $\uparrow$ & DINO-I $\uparrow$ & DINO-C $\uparrow$ & F1 score $\uparrow$ \\
 & & & & 0.410 & 0.230 & 0.593 & 0.610 & 0.361 \\
\checkmark & & & & 0.411 & 0.225 & 0.591 & 0.637 & 0.364 \\
\checkmark & \checkmark & & & \textbf{0.407} & 0.228 & 0.600 & 0.635 & 0.653 \\
\checkmark & \checkmark & \checkmark & & 0.408 & 0.231 & \textbf{0.618} & 0.648 & 0.718 \\
\midrule
\checkmark & \checkmark & \checkmark & \checkmark & \textbf{0.407} & \textbf{0.235} & \textbf{0.618} & \textbf{0.651} & \textbf{0.727} \\
\bottomrule
     \end{tabular}}
\end{table}

\cref{tab:exp-ablation} presents the quantitative ablation study of DiffSensei, where we systematically remove components to assess their impact. Specifically, when the MLLM component, serving as the flexible character feature adapter, is excluded, the CLIP metrics decrease by 1.73\%, and the DINO-C score also drops, underscoring its role in enhancing transferring character to the text-derived states. The absence of the Magi~\cite{magi} image encoder results in a general decline in metrics, particularly in image quality and character similarity, highlighting the importance of the Magi encoder for effectively encoding manga characters. Magi is trained specifically on manga datasets, performing better at preserving manga characters.
To investigate alternative methods for encoding character and dialog layout conditions, we experimented with replacing the dialog embedding technique by inputting the Fourier embeddings of dialog bounding boxes into the timestep embedding of SDXL~\cite{sdxl}. This modification led to a significant decrease in layout control, evidenced by the F1 score plummeting from 0.653 to 0.364, demonstrating that directly incorporating dialog embeddings into the latent is a superior approach for encoding dialog layouts.
Furthermore, we explored adding the Fourier embeddings of character bounding boxes to the character features as an alternative to masked attention injection. This change caused a marked drop in the DINO-C metric, reaffirming the effectiveness of our original masked attention strategy. For comprehensive effect showcases, please see the qualitative ablation study in the Appendix.
\section{Conclusion}
\label{sec:conclusion}

This paper introduces DiffSensei, a novel framework for multi-character customized story visualization that integrates a diffusion-based image generator with an MLLM as a text-compatible identity adapter. Key innovations include masked attention control for character layout management, dialog layout embedding, and an MLLM-based feature adapter for flexible character customization. Supported by the proposed MangaZero dataset, comprising 43,264 manga pages and 427,147 panels, DiffSensei achieves superior, character-consistent panels that dynamically respond to textual prompts, surpassing existing methods and advancing the field of story visualization.

\noindent
\textbf{Acknowledgement.} This work is supported by the National Key Research and Development Program of China (No. 2023YFC3807600).

{
    \small
    \bibliographystyle{ieee_fullname}
    \bibliography{egbib}
}

\appendix

\twocolumn[{
\centering
\section*{DiffSensei: Bridging Multi-Modal LLMs and Diffusion Models \\ for Customized Manga Generation \\ \textnormal{Supplementary Material}}
\vspace{5mm} 
}]

\noindent
\textbf{Overview.}
\begin{itemize}
    \item \textbf{\cref{sec:supp-more-qualitative-results}.} More qualitative results.
    \item \textbf{\cref{sec:supp-compare}.} More qualitative comparison results.
    \item \textbf{\cref{sec:supp-implementation-details}.} Implementation details of the experiments.
    \item \textbf{\cref{sec:supp-ablation-study}.} More ablation studies.
    \item \textbf{\cref{sec:supp-limitations}.} Limitations and failure cases of our model.
    \item \textbf{\cref{sec:supp-dataset-details}.} Details of the MangaZero dataset.
    \item \textbf{\cref{sec:supp-broader-impacts}.} Broader impacts.
\end{itemize}


\section{More Qualitative results}

\begin{figure*}[t!]
	\centering
	\includegraphics[width=1.0\linewidth]{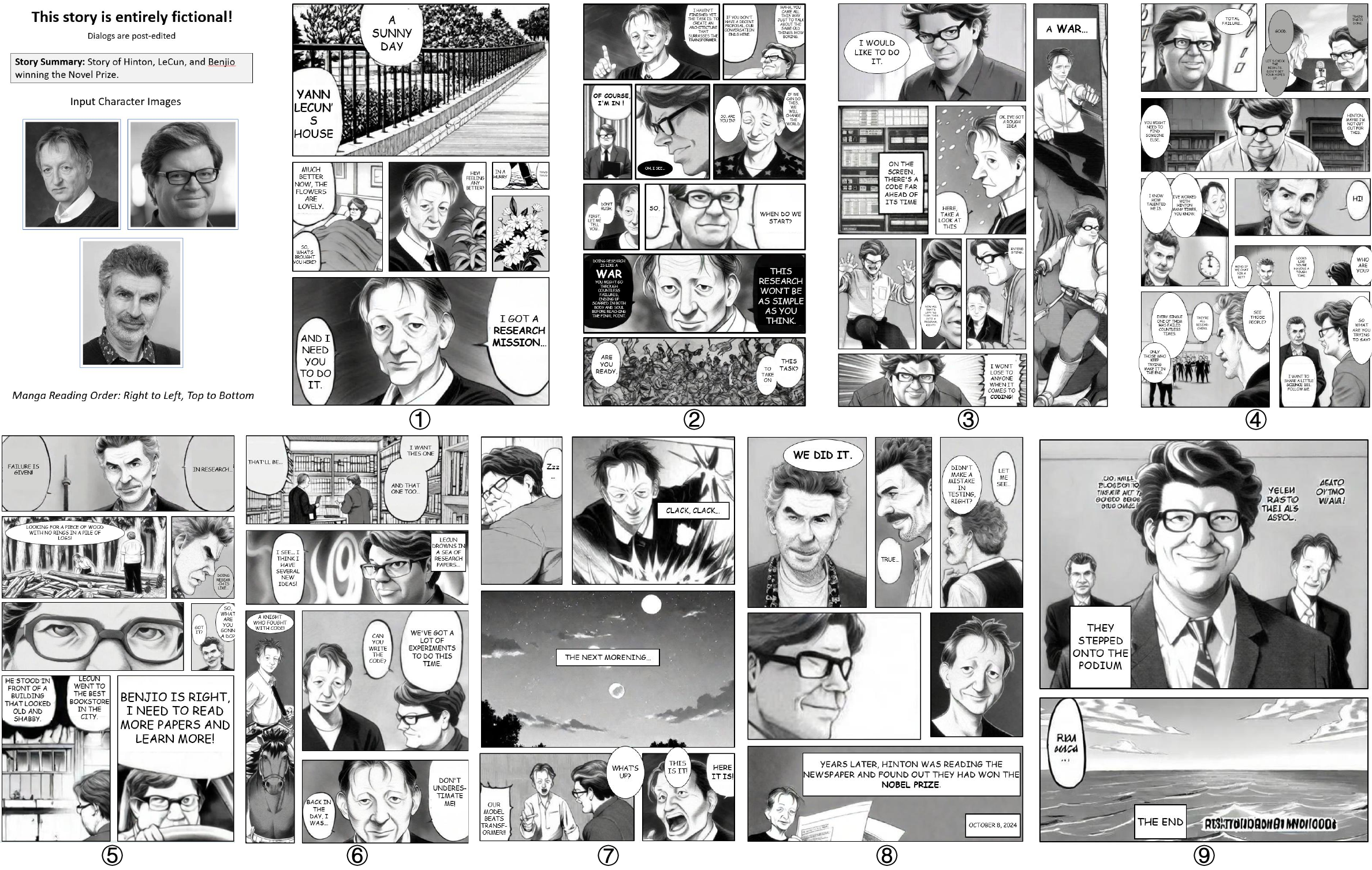}
	\caption{A complete long manga story about Hinton, LeCun, and Bengio winning the Nobel Prize.}
	\label{fig:supp-nobel-prize}
\end{figure*}

\label{sec:supp-more-qualitative-results}
Due to the extensive qualitative results, we have presented them in separate PDF files. Please refer to the \href{https://jianzongwu.github.io/projects/diffsensei/}{project page} for more details. We summarize the content of the documents below:

``\textbf{page\_results.pdf}'' presents additional manga page examples similar to those in Fig. 6 of the main paper. The numerous examples demonstrate that our DiffSensei model can generate vivid manga pages featuring diverse storylines, characters, and backgrounds. Notably, when provided with previously unseen character images, DiffSensei also performs well, highlighting the model's generalization capabilities. Some illustrations are also shown in~\cref{fig:supp-page-caption}, \cref{fig:supp-page-caption2}, \cref{fig:supp-page-results1}, and~\cref{fig:supp-page-results2}.

``\textbf{long\_story.pdf}'' showcases a complete, extended manga story about Hinton, LeCun, and Bengio winning the Nobel Prize—an expansion of the real human manga shown in Fig. 1. The manga tells a fictional story of researchers Hinton, LeCun, and Bengio taking on a challenge to create an AI model surpassing Transformers. Facing failures and self-doubt, they persist through rigorous research and collaboration. After overcoming numerous hurdles, their model succeeds, and years later, they are awarded the Nobel Prize, celebrating their groundbreaking achievement and the power of perseverance in science. This full version illustrates that our model can create comprehensive long stories in a zero-shot setting, effectively handling real human-centric manga narratives. This story is also shown in~\cref{fig:supp-nobel-prize}.

\section{More Qualitative Comparison Results}
\label{sec:supp-compare}

\begin{figure*}[t!]
	\centering
	\includegraphics[width=1.0\linewidth]{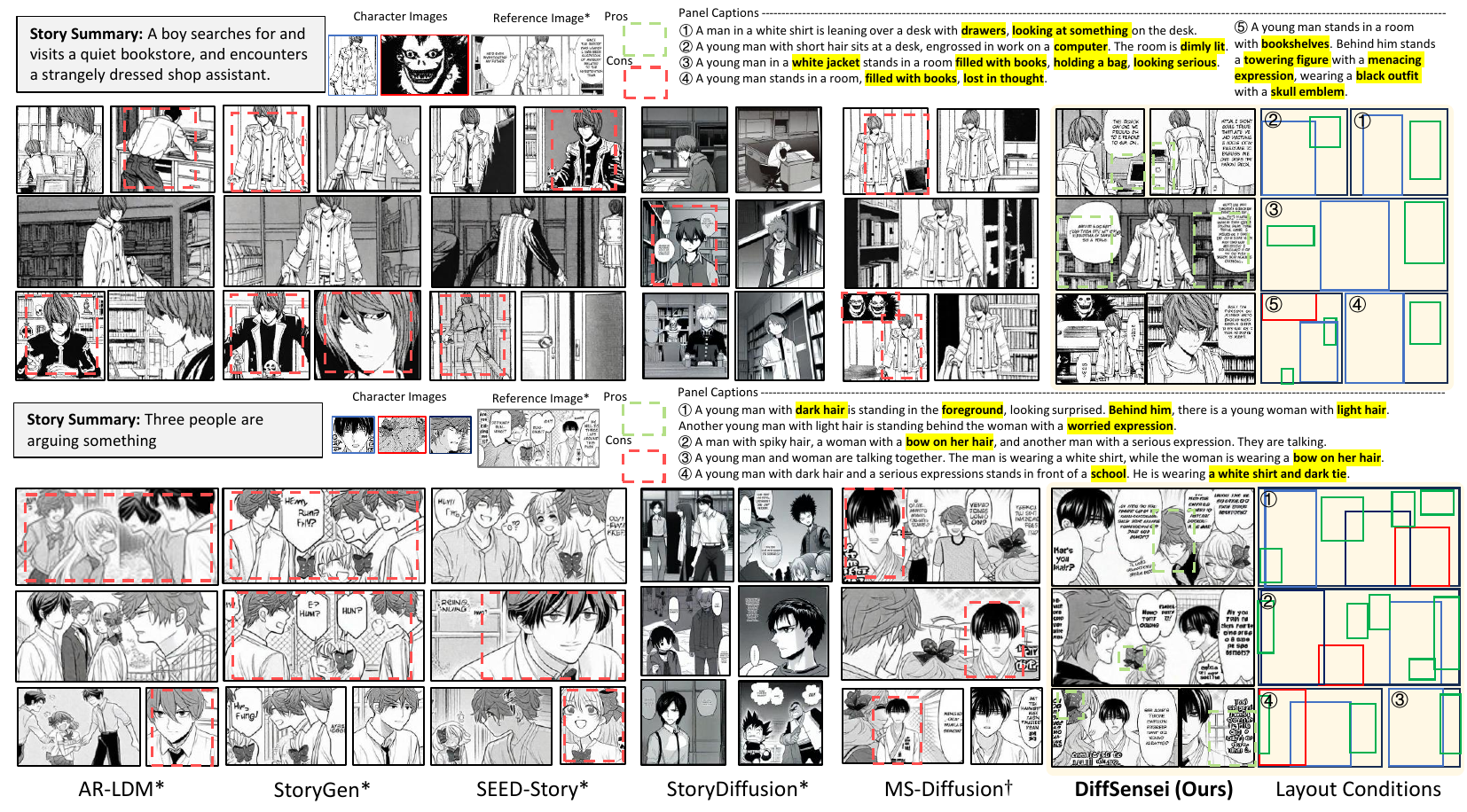}
	\caption{\textbf{More qualitative comparisons with baselines}. Baselines followed by a ``*'' use reference images as input rather than character images. Methods marked by ``\dag'' means re-trained with dialog embedding.}
	\label{fig:supp-comparison}
\end{figure*}

\begin{figure}[t!]
	\centering
	\includegraphics[width=1.0\linewidth]{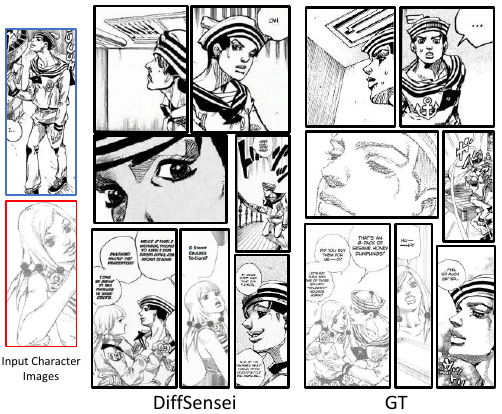}
	\caption{\textbf{Comparison with ground truth}. The results generated by DiffSensei show remarkable consistency and aesthetics compared to ground truth manga panels.}
	\label{fig:supp-gt-comparison}
\end{figure}

The additional qualitative comparisons between DiffSensei and baseline models are presented in~\cref{fig:supp-comparison}. 
The observations align closely with those discussed in the main paper. Models such as AR-LDM~\cite{ar-ldm} and StoryGen~\cite{storygen} cannot process separate character images as inputs, limiting their ability to control the layout of individual characters explicitly. 
SEED-Story~\cite{seed-story} incorporates an MLLM to predict panel captions, but its effectiveness is constrained, resulting in some unnatural storytelling outcomes. 
StoryDiffusion~\cite{storydiffusion} employs a training-free approach, which underperforms likely due to its inability to effectively handle varying-resolution reference image inputs. 
MS-Diffusion~\cite{ms-diffusion} demonstrates comparatively better performance in identity preservation and character layout control. 
However, it tends to constrain character input images overly, lacking the flexibility to adjust character appearances based on textual inputs dynamically. 
In contrast, our DiffSensei model outperforms all baselines, achieving superior results on \textit{identity preservation, text compatibility, and overall image quality}.

Moreover, we show the comparison with ground truth in~\cref{fig:supp-gt-comparison}. The result highlights the remarkable visual quality and character consistency of DiffSensei. DiffSensei achieves close or even better character image alignment than ground truth.

\section{Implementation Details}
\label{sec:supp-implementation-details}

\noindent
\textbf{Inference details.} During inference, we follow prior works~\cite{ip-adapter, ms-diffusion} by setting the character feature weight to $\alpha = 0.6$. 
Additionally, we weight the MLLM-adapted character features according to $\mathbf{c}'_i = (1 - \beta)\mathbf{c}_i + \beta \hat{\mathbf{c}}_i$. In all our experiments, we set $\beta = 0.4$, which provides an effective balance between identity preservation and text compatibility.

\noindent
\textbf{Training details.}
Our model training is conducted on 8 NVIDIA A100 80G GPUs, requiring approximately one week for the first stage and one day for the second. Before beginning the first stage, we carry out a pre-training phase where we fine-tune the SDXL model~\cite{sdxl} on manga images with the text-to-image task. This pre-training helps the model adapt to the black-and-white manga distribution, accelerating subsequent training. 
The pre-training required only 10k steps and can be completed within 5 hours.

\noindent
\textbf{Human evaluation details.}

\begin{figure}[t!]
	\centering
	\includegraphics[width=1.0\linewidth]{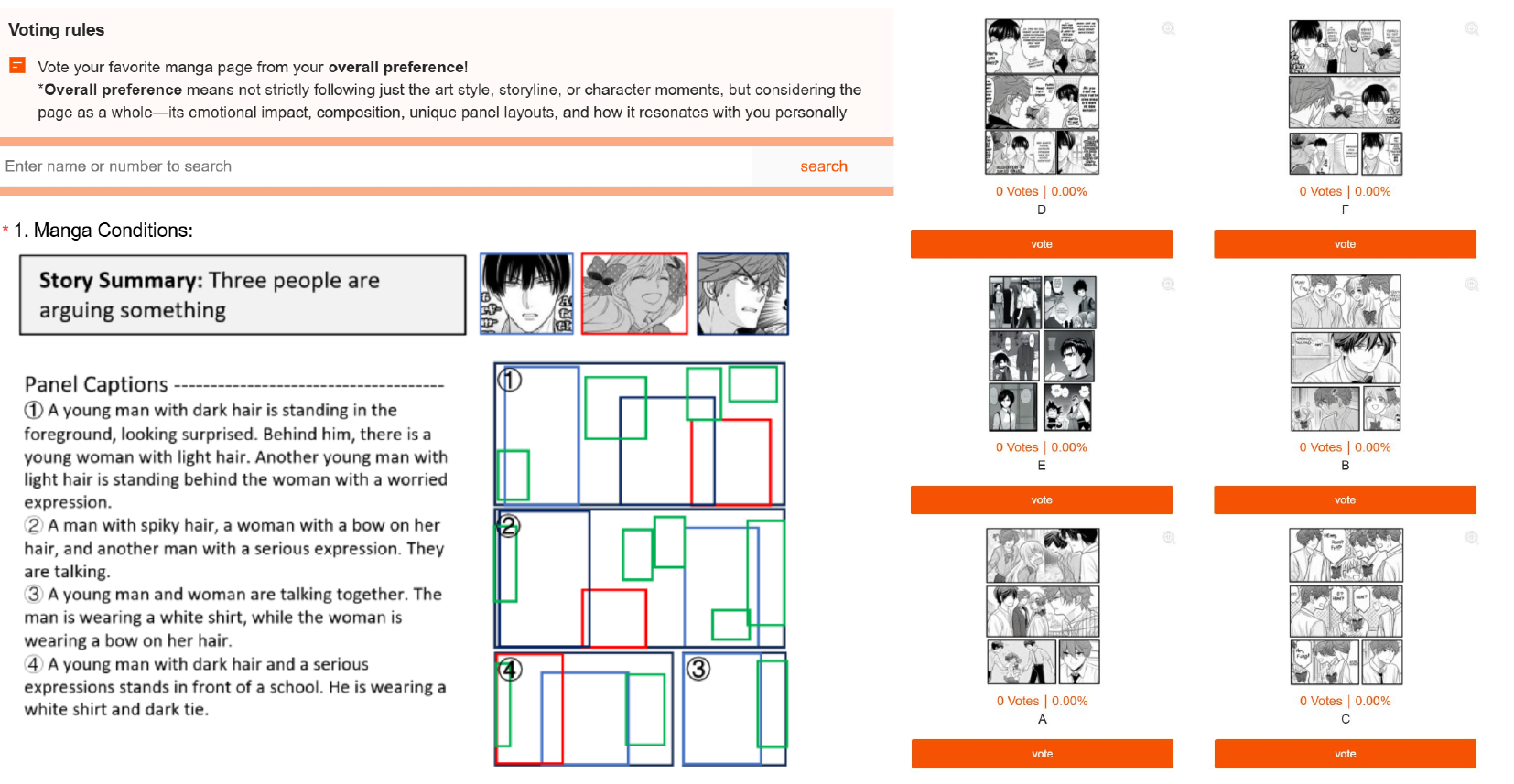}
	\caption{\textbf{Human evaluation interface.}}
	\label{fig:supp-human-evaluation-foreground}
\end{figure}

For human evaluation, we invited 15 general users from diverse backgrounds, ages, and genders to assess 10 anonymized manga page sets each. Evaluators reviewed the pages from different aspects, and their ratings were averaged to compute the preference probability across methods. The interface is shown in~\cref{fig:supp-human-evaluation-foreground}.

\section{Ablation Study}
\label{sec:supp-ablation-study}

\begin{figure}[t!]
	\centering
	\includegraphics[width=1.0\linewidth]{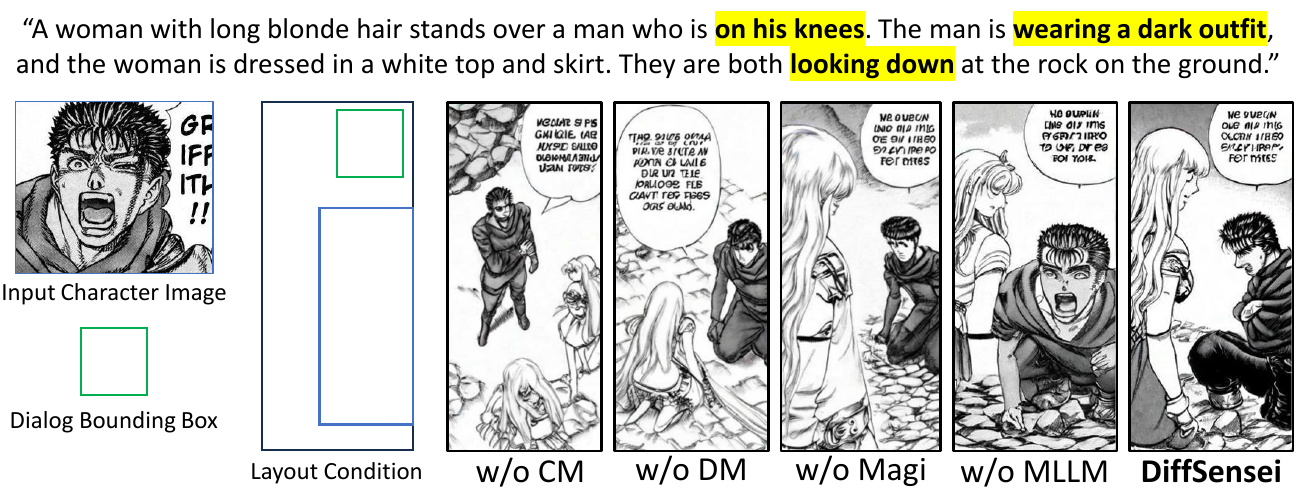}
	\caption{\textbf{Qualitative ablation of the proposed modules}. CM is character masked attention injection. DM is dialog masked encoding. Magi means using Magi~\cite{magi} image encoder. MLLM means using MLLM for stage 2 training.}
	\label{fig:supp-ablation-module}
\end{figure}

\begin{figure}[t!]
	\centering
        \includegraphics[width=1.0\linewidth]{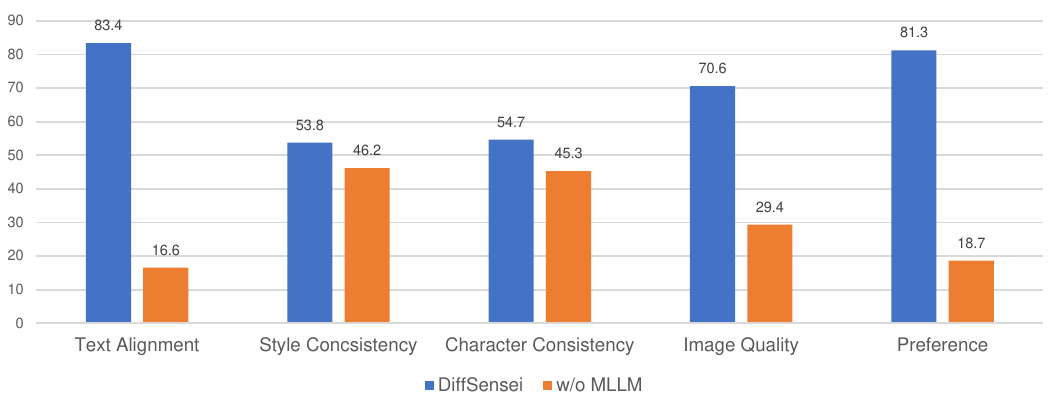}
	\caption{\textbf{Human evaluation on the MLLM module.}}
	\label{fig:supp-mllm-human-preference}
\end{figure}

\begin{table*}[!ht]
    \centering
    \caption{\textbf{Quantitative ablations}. The first scores are \textbf{bold}. The second scores are \underline{underlined}.}
    \begin{subtable}{0.49\linewidth}
        \centering
        \caption{Ablation on the rate of character image sourced from target panel.}
        \label{tab:supp-ablation-self-rate}
        \scalebox{0.82}{\begin{tabular}{l|ccccc}
            \toprule
            Rate & FID $\downarrow$ & CLIP $\uparrow$ & DINO-I $\uparrow$ & DINO-C $\uparrow$ & F1 score $\uparrow$ \\
            \midrule
            0.0 & 0.408 & 0.233 & 0.615 & 0.646 & 0.718 \\
            0.5 & \textbf{0.407} & \textbf{0.235} & \textbf{0.618} & \textbf{0.651} & 0.727 \\
            1.0 & \textbf{0.407} & 0.233 & 0.610 & 0.644 & \textbf{0.729} \\
            \bottomrule
        \end{tabular}}
    \end{subtable}
    \begin{subtable}{0.49\linewidth}
        \centering
        \caption{Ablation on the MLLM weighting hyperparameter $\beta$.}
        \label{tab:supp-ablation-beta}
        \scalebox{0.72}{\begin{tabular}{l|ccccc}
            \toprule
            $\beta$ & FID $\downarrow$ & CLIP $\uparrow$ & DINO-I $\uparrow$ & DINO-C $\uparrow$ & F1 score $\uparrow$ \\
            \midrule
            0.0 & 0.408 & 0.231 & \underline{0.618} & 0.648 & 0.718 \\
            0.2 & \underline{0.407} & 0.231 & \textbf{0.620} & \textbf{0.653} & 0.722 \\
            0.4 & \underline{0.407} & 0.235 & \underline{0.618} & \underline{0.651} & \underline{0.727} \\
            0.6 & \textbf{0.406} & \textbf{0.237} & 0.608 & 0.637 & 0.728 \\
            0.8 & \underline{0.407} & \textbf{0.237} & 0.604 & 0.629 & 0.727 \\
            1.0 & \underline{0.407} & \underline{0.236} & 0.601 & 0.618 & \textbf{0.731} \\
            \bottomrule
        \end{tabular}}
    \end{subtable}
\end{table*}

\begin{figure*}[t!]
	\centering
	\includegraphics[width=1.0\linewidth]{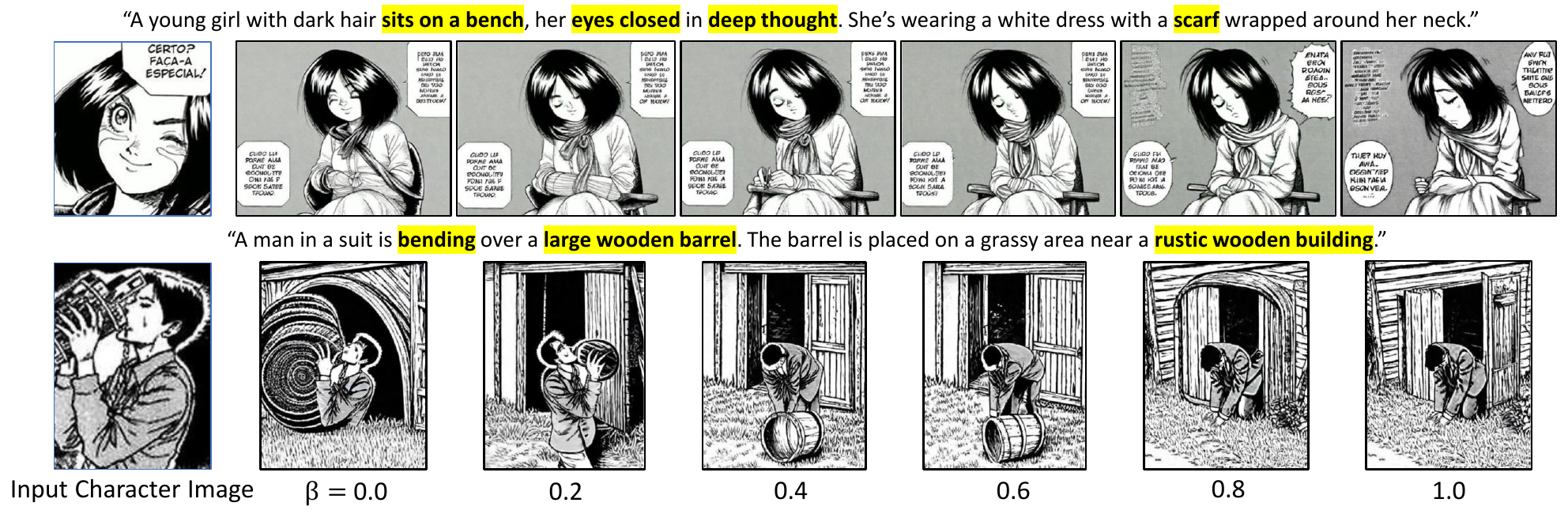}
	\caption{\textbf{Qualitative ablation of $\beta$}.}
	\label{fig:supp-ablation-beta}
\end{figure*}

\noindent
\textbf{Qualitative ablation about the proposed modules.}
\cref{fig:supp-ablation-module} presents a qualitative ablation study of the proposed modules. Without the character masked attention injection, the character layout cannot be effectively controlled. Similarly, replacing the masked embedding technology with Fourier embedding for dialog layout control results in incorrect dialog generation. 
The Magi~\cite{magi} image encoder is explicitly trained on manga images. When it is not used, and only the CLIP image encoder is employed, the ability to preserve character identity is significantly degraded.
Additionally, without using the MLLM as a flexible character adapter, the model tends to rigidly adhere to the input character image's pixel distribution, which limits its ability to adapt character appearances, poses, and motions according to the text input. 
Specifically, the man is not kneeling or looking down on the ground.

\noindent
\textbf{Rate of character image sourced from target panel.}
Thanks to the character annotations, we can identify the same character across multiple panels, which allows us to capture different appearances, poses, and motions. During training, we sample a source character using its character ID, which remains consistent throughout the page. 
However, we find that using a completely random sample made training difficult to converge, likely due to the artistic exaggerations often found in manga. 
These exaggerations make it challenging for the model to learn consistent character representations when the character's appearance changes significantly across panels.
To address this, we introduced an alternative sampling strategy where, at a set rate, the source character is sampled directly from the target image itself. This helps the model learn identity preservation more effectively. An ablation study examining the impact of the target character sampling rate is presented in~\cref{tab:supp-ablation-self-rate}. 
The results indicate that a higher sampling rate decreases text compatibility, as the model becomes overly focused on replicating the input character image. 
On the other hand, a lower sampling rate makes it difficult to converge the training, which also negatively impacts the metrics. Ultimately, we choose a sampling rate of 0.5 during training to provide a balanced performance.

\noindent
\textbf{Human evaluation of the MLLM module.}
We conducted a human evaluation on the MLLM module. The results are presented in~\cref{fig:supp-mllm-human-preference}. The full model with MLLM demonstrates a significant advantage, particularly in terms of text alignment and overall preference. The results indicate that the MLLM module enhances both the visual appeal and text compatibility of manga generation.

\noindent
\textbf{Hyperparameter $\beta$.}
\cref{tab:supp-ablation-beta} presents a quantitative ablation study of the MLLM weighting hyperparameter, $\beta$. 
The results suggest that increasing $\beta$ enhances the CLIP score but decreases the DINO score. 
This demonstrates that while the MLLM-adapted embeddings become more aligned with text, they may compromise identity preservation. 
Conversely, a lower $\beta$ decreases the CLIP score, favoring better identity retention. Empirically, we set $\beta$ to 0.4 to achieve an optimal balance between these factors. Qualitative results in~\cref{fig:supp-ablation-beta} further support this observation. As $\beta$ increases, character preservation diminishes. Conversely, with smaller values of $\beta$, the generated character lacks compatibility with the text. In this example, the generated images achieve the best balance when $\beta$ is set to 0.4 or 0.6.

\section{Limitations and Future Work}
\label{sec:supp-limitations}

\begin{figure}[t!]
	\centering
	\includegraphics[width=1.0\linewidth]{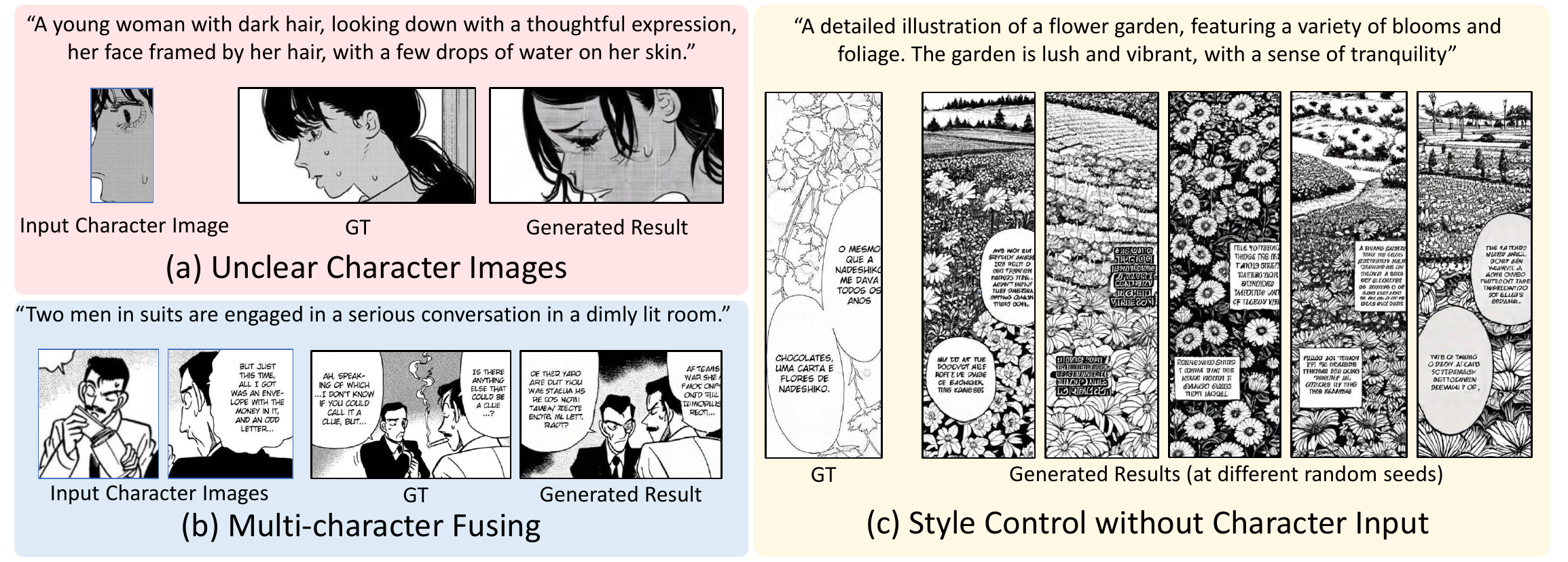}
	\caption{\textbf{Failure cases}.}
	\label{fig:supp-failure-cases}
\end{figure}

\begin{figure*}[!ht]
    \centering
    \centering
    \begin{subfigure}{1.0\textwidth}
        \includegraphics[width=\textwidth]{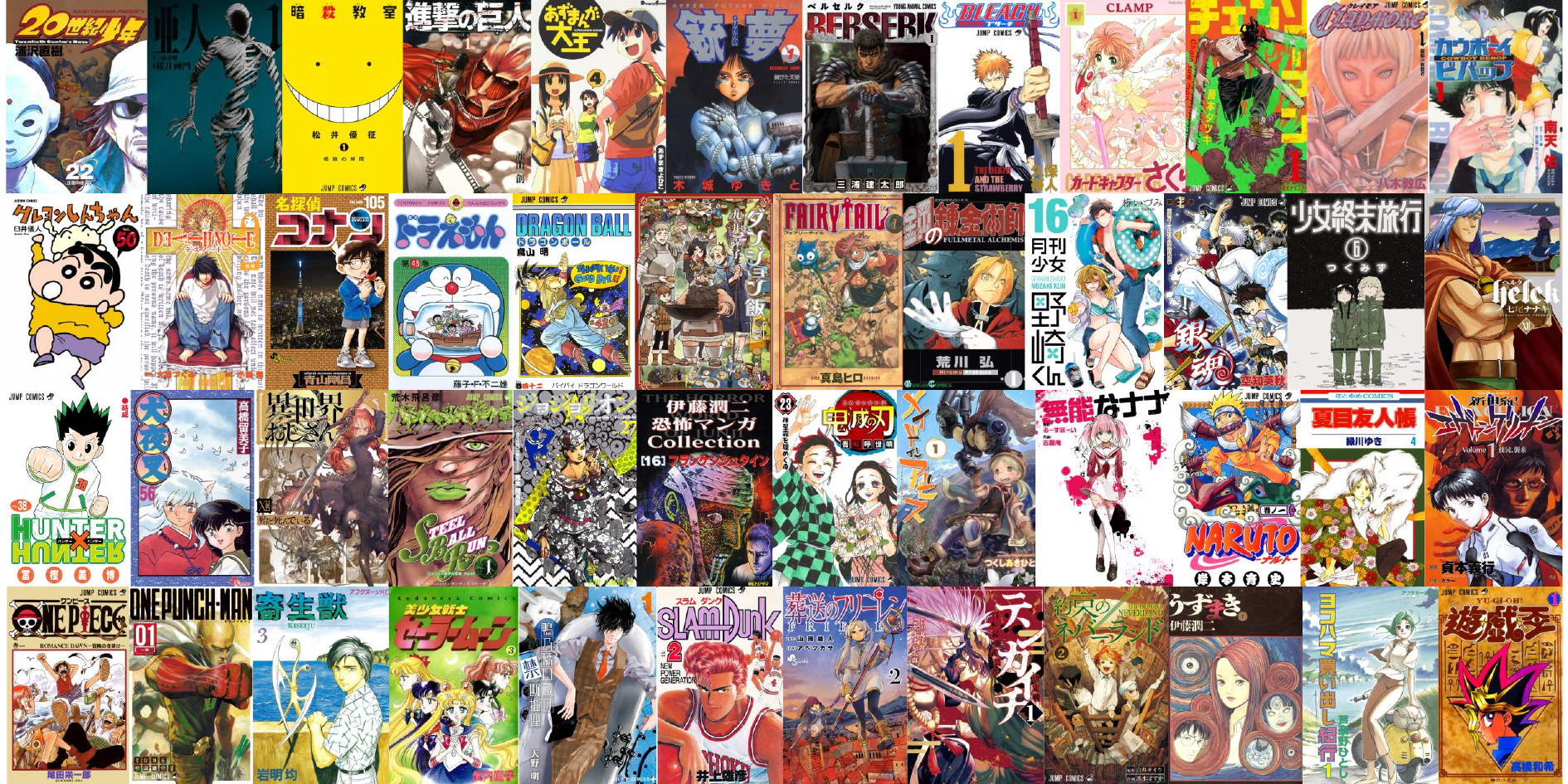}
        \caption{Covers of manga series.}
        \label{fig:supp-dataset-cover}
    \end{subfigure}
    \begin{subfigure}{0.77\textwidth}
        \includegraphics[width=\textwidth]{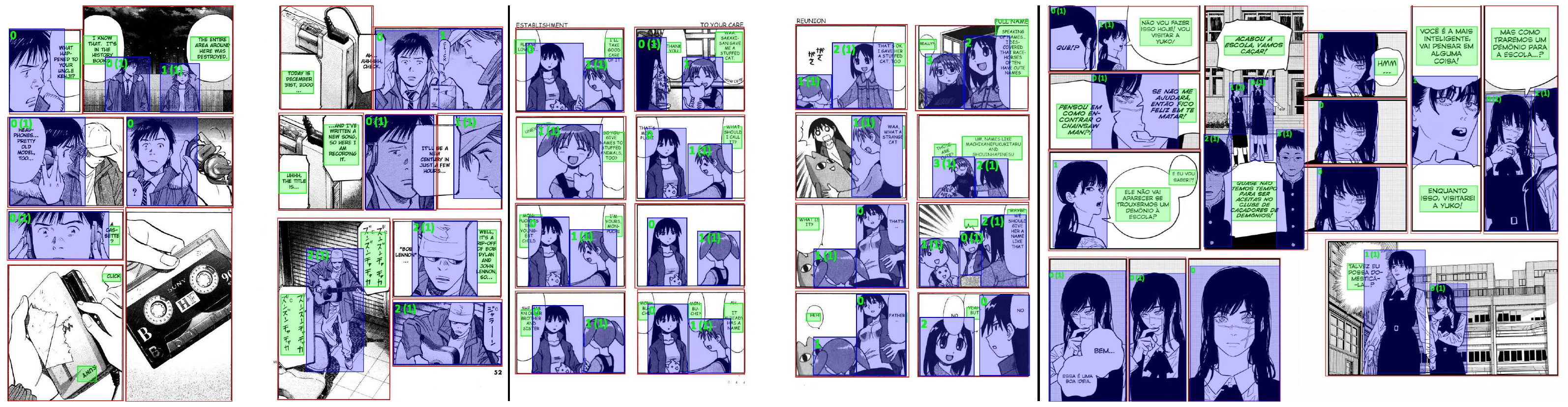}
        \caption{Examples of character and dialog annotations.}
        \label{fig:supp-dataset-annotation}
    \end{subfigure}
    \begin{subfigure}{0.22\textwidth}
        \includegraphics[width=\textwidth]{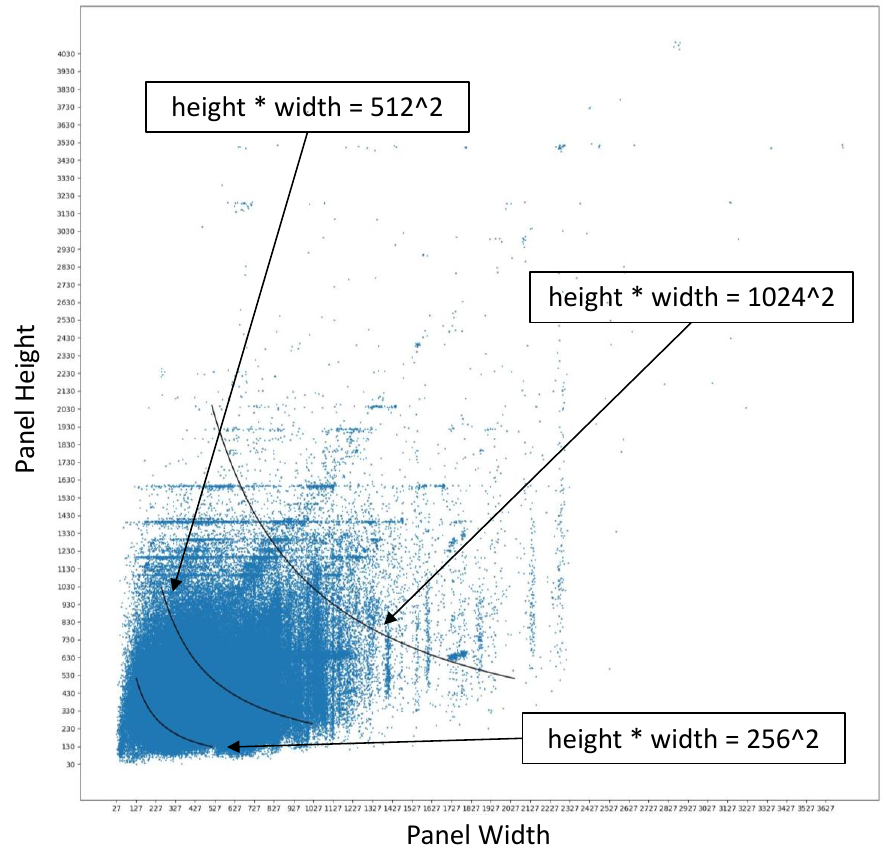}
        \caption{Resolution distribution.}
        \label{fig:supp-dataset-resolution-distribution}
    \end{subfigure}
    \caption{\textbf{Details of the MangaZero dataset}.}
\end{figure*}

\cref{fig:supp-failure-cases} showcases several failure cases of DiffSensei.
(a) \textbf{Unclear input character images}: When the input character image is vague or unclear, our model struggles to capture the character's explicit appearance, resulting in a loss of identity. This issue could be mitigated by refining the dataset or restricting user inputs to ensure that the character images are sufficiently clear for accurate reproduction.
(b) \textbf{Multiple character fusing}. When multiple characters are provided as input, our model sometimes exhibits a ``fusing phenomenon,'' where the characters appear more similar than they actually are, especially when their original appearances are already quite alike. This is likely influenced by limitations of the base model (SDXL), which also demonstrates this problem~\cite{sdxl}. Future work could investigate more advanced text-to-image models or develop methods that better disentangle multiple character representations.
(c) \textbf{Style control without character input}. When generating manga panels without any character inputs, the model struggles to control the output style. While the ground truth consists of simple line drawings with a distinct style, the generated images tend to adopt a more generic manga appearance, failing to achieve precise style control. Notably, when character images are provided, our approach can control the overall image style to match that of the character. Future research could focus on improving style control for different manga styles, especially in scenarios without character inputs.

\section{MangaZero Dataset Details}
\label{sec:supp-dataset-details}

\noindent
\textbf{Manga sources.}
The MangaZero dataset contains the most famous Japanese black-and-white manga series. \cref{fig:supp-dataset-cover} shows the cover pages of all the 48 series. These manga series were selected primarily for their popularity, distinct art style, and extensive casts of characters, providing our model with the capacity to develop a robust and flexible ability for identity preservation.

\noindent
\textbf{Annotations.}
Examples of character and dialogue annotations are illustrated in \cref{fig:supp-dataset-annotation}. Additionally, \cref{fig:supp-dataset-resolution-distribution} depicts the panel resolution distribution within the dataset. To enhance clarity, we include three reference lines representing resolutions of 1024$\times$1024, 512$\times$512, and 256$\times$256. Most manga panels cluster around the second and third lines, indicating that most panels are of relatively low resolution compared to those typically emphasized in recent studies~\cite{sdxl, flux}. This characteristic is inherent to manga data, which our work specifically addresses. Consequently, multi-resolution training becomes essential for effectively working with manga datasets.

\noindent
\textbf{Potential usage and future work.}
Our dataset is designed primarily for the task of customized manga generation, offering substantial versatility for further applications. Beyond its core purpose, it can be utilized in other promising research areas. For instance, it is well-suited for customized manga continuation, where the goal is to generate coherent story extensions based on an initial panel or sequence. This task leverages the intrinsic reading order of manga, allowing panels to be organized in a narrative flow, making it possible for models to autonomously extend a storyline while maintaining visual and thematic consistency.
Additionally, by expanding the scope of annotations, our dataset could foster research into style control for manga generation. Each manga series or artist possesses a distinctive drawing style, and our dataset is well-positioned to support training models in style controllability. This would enable fine-tuning or conditional generation of manga that can mimic specific artistic styles, providing a nuanced tool for researchers exploring stylistic variation and artist-specific customization in manga creation.
Overall, the dataset's adaptability and extensibility make it an excellent foundation for future research into both narrative continuation and stylistic control, promising avenues for extending the capabilities of manga generation models.

\section{Broader Impacts}
\label{sec:supp-broader-impacts}

The broader impacts of this paper are significant across several domains.

\noindent
\textbf{Manga industry.}
The proposed technology can directly benefit the manga industry by enabling artists, creators, and publishing houses to rapidly create customized manga with detailed control over characters and layouts. 
This innovation can potentially \textit{streamline} the manga production process, reduce production costs, and allow for more personalized storytelling that caters to niche audiences or specific market demands.

\noindent
\textbf{Education, film, and media.}
Beyond manga, the framework’s ability to visualize stories from text could be impactful in fields like education, film, and media production. In educational settings, it can assist in generating visual aids that align with narratives, enhancing engagement and comprehension for students. In filmmaking, it could serve as a pre-visualization tool to quickly draft visual storyboards, facilitating ideation and communication between writers, directors, and production teams.

\noindent
\textbf{Ethical use of data.}
The approach emphasizes the importance of legal data usage, particularly stressing the licensing and ethical constraints of the MangaZero dataset or more manga data annotated like MangaZero. This sets a precedent for responsible data handling in the domain, ensuring that generated content respects copyright laws and that the data is either properly licensed or restricted to academic and non-commercial usage. This awareness mitigates potential legal issues and supports ethical AI research practices in creative fields.

\begin{figure*}[t!]
	\centering
	\includegraphics[width=1.0\linewidth]{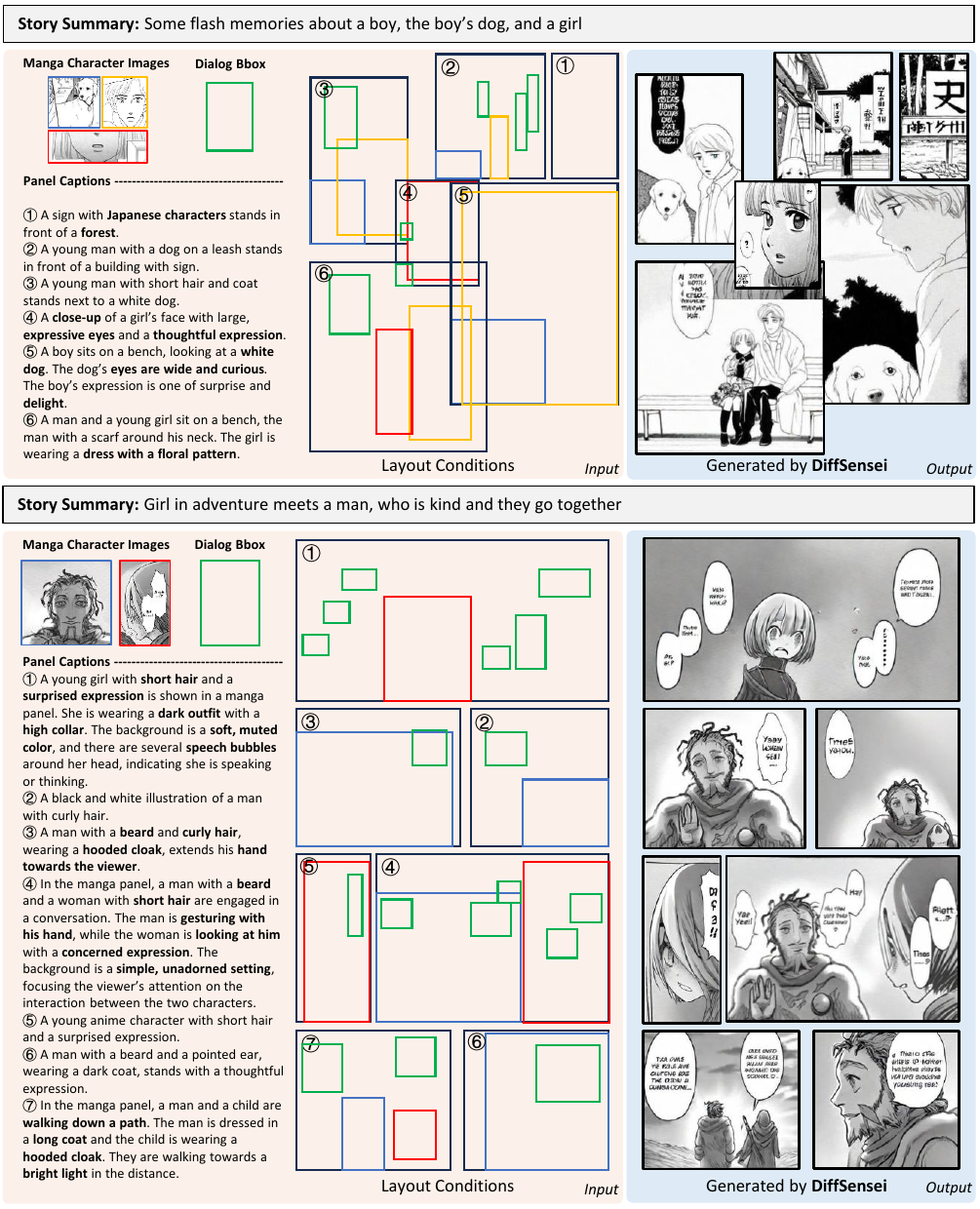}
	\caption{DiffSensei generated results with inputs (Part1).}
	\label{fig:supp-page-caption}
\end{figure*}

\begin{figure*}[t!]
	\centering
	\includegraphics[width=1.0\linewidth]{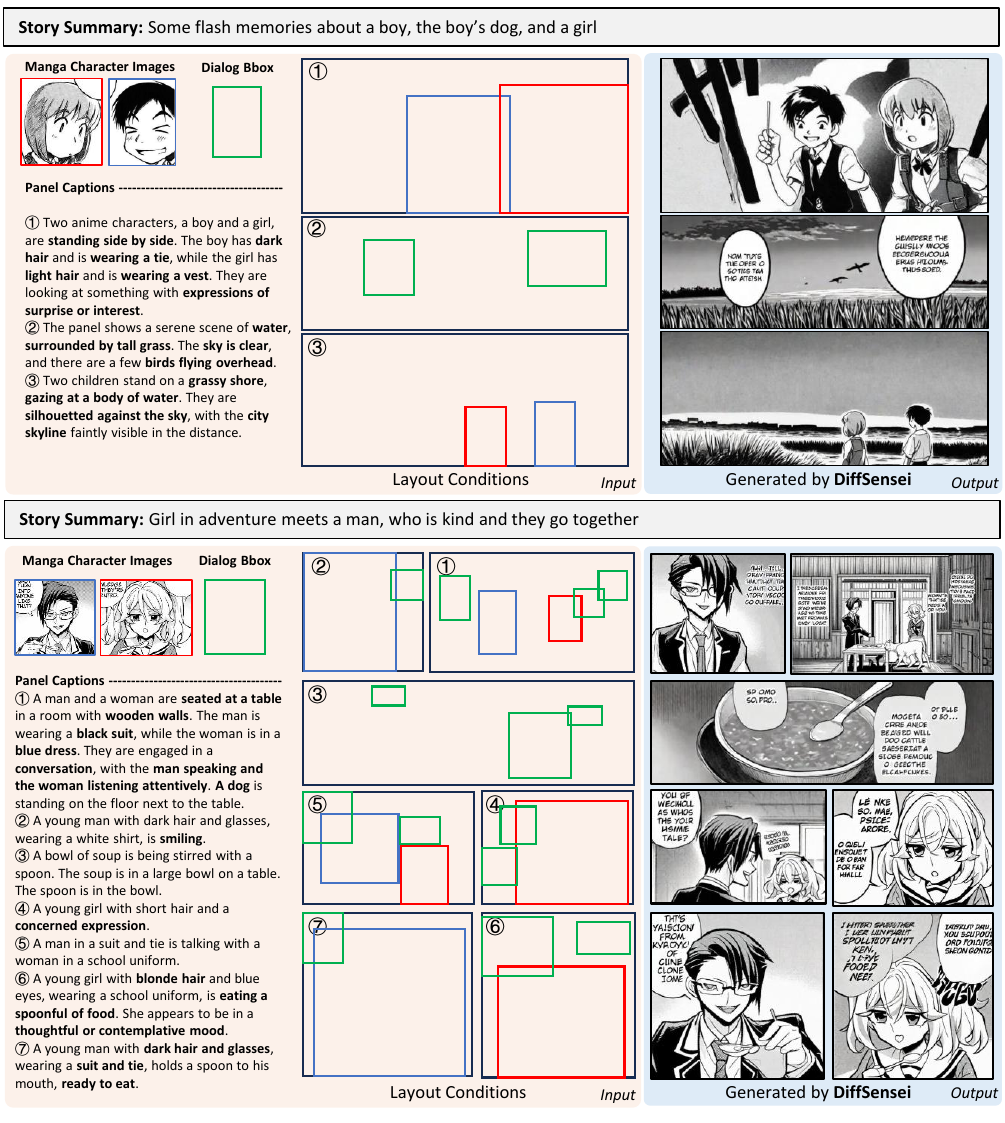}
	\caption{DiffSensei generated results with inputs (Part2).}
	\label{fig:supp-page-caption2}
\end{figure*}

\begin{figure*}[t!]
	\centering
	\includegraphics[width=1.0\linewidth]{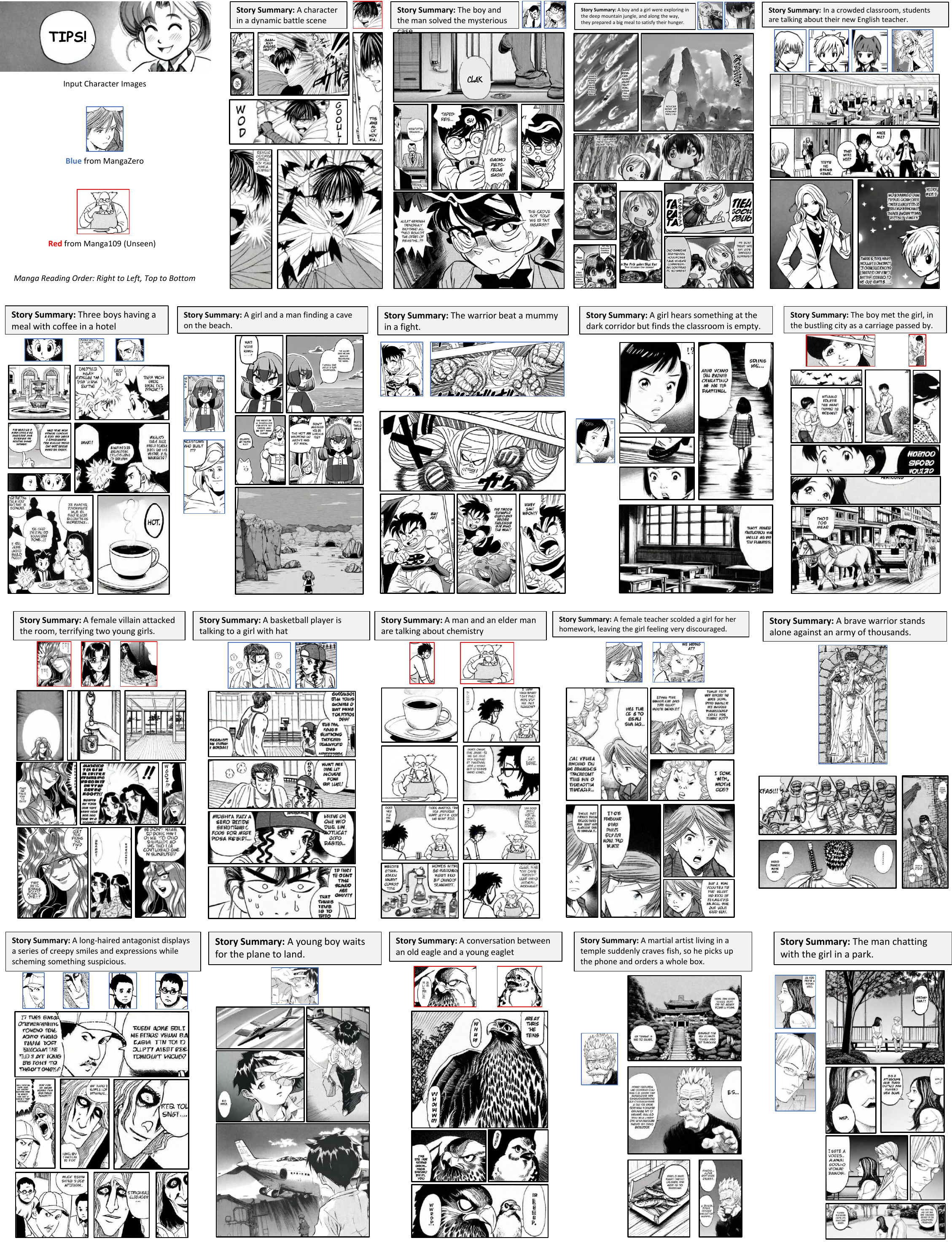}
	\caption{Manga pages generated by DiffSensei (Part 1).}
	\label{fig:supp-page-results1}
\end{figure*}

\begin{figure*}[t!]
	\centering
	\includegraphics[width=1.0\linewidth]{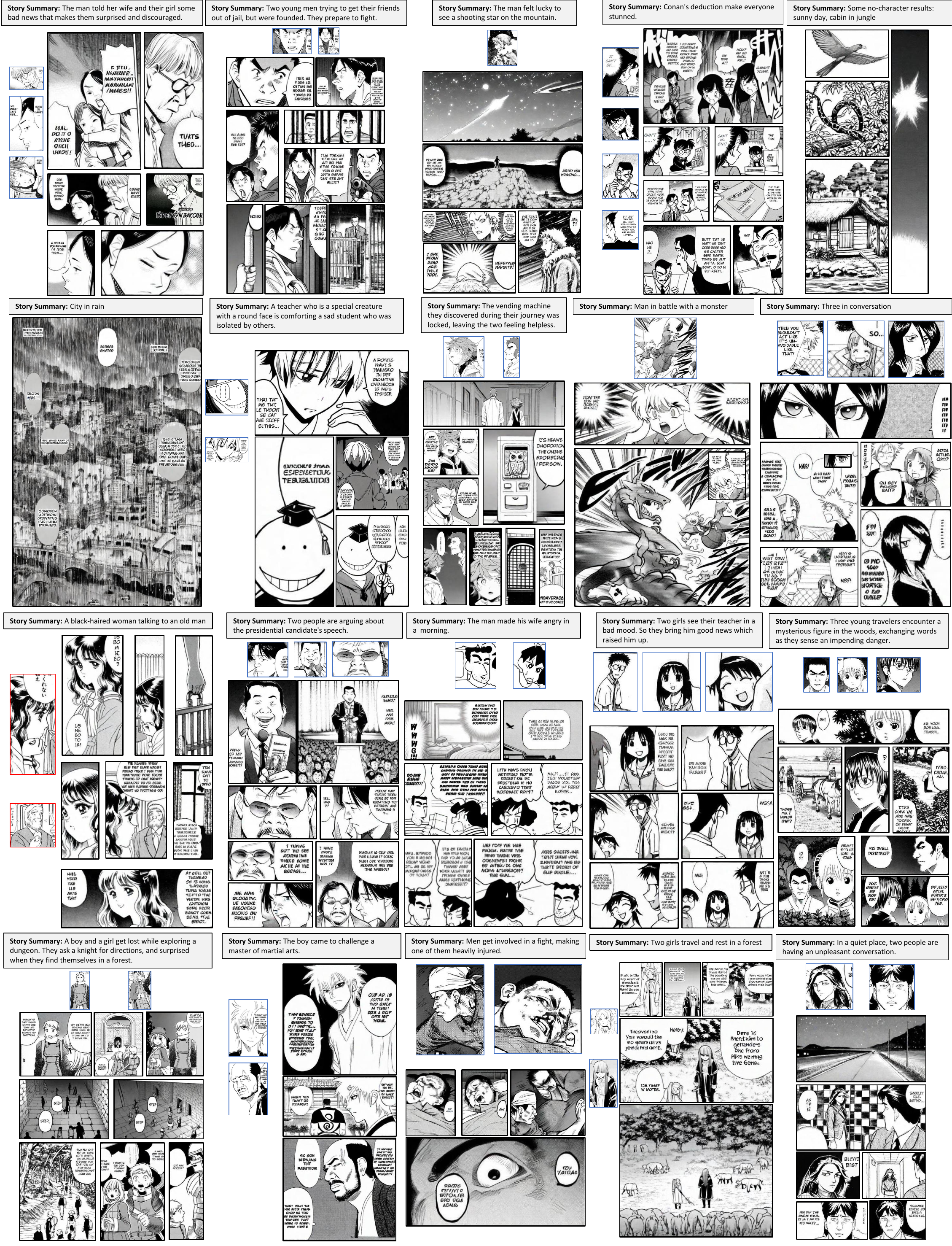}
	\caption{Manga pages generated by DiffSensei (Part 2).}
	\label{fig:supp-page-results2}
\end{figure*}

\end{document}